\documentclass[10pt]{article}

\usepackage{arxiv}

\usepackage{epstopdf}

\usepackage[T1]{fontenc}
\usepackage{newtxtext}
\usepackage{amsmath}
\usepackage{amsthm}
\usepackage{amsfonts}
\usepackage{amssymb}
\usepackage[nolist]{acronym}
\usepackage[utf8]{inputenc}
\usepackage{textcomp}
\usepackage{pifont}
\newcommand{\cmark}{\ding{51}}%
\newcommand{\xmark}{\ding{55}}%

\usepackage{listings}
\usepackage{matlab-prettifier}
\usepackage{multicol}
\usepackage{subcaption}
\usepackage{graphicx}  
\usepackage{xcolor}
\usepackage{float}
\usepackage{csquotes}

\usepackage{siunitx}
\usepackage{xspace}

\usepackage[ruled]{algorithm2e}

\usepackage{color}
\usepackage{mathtools}

\usepackage{multirow}

\usepackage{eurosym} 

\usepackage{nicefrac}

\usepackage{smartref}
\usepackage{tikz}
\usepackage{pgfplots}
\pgfplotsset{compat=1.18}
\usepackage{url}

\usetikzlibrary{arrows,shapes,positioning,calc,intersections,through,backgrounds,fit,decorations.markings, arrows.meta, datavisualization}
\usepackage{csvsimple}
\usetikzlibrary{spy}
\usepackage{adjustbox}
\usetikzlibrary{backgrounds}
\usepgfplotslibrary{groupplots}
\usepgfplotslibrary{statistics}

\usepackage{ifplatform}
\usepackage{tikz-3dplot}
\usepgfplotslibrary{patchplots} 

\usetikzlibrary{arrows.meta, bending}
\usepackage{ifthen}

\tikzset{
	partial ellipse/.style args={#1:#2:#3}{
		insert path={+ (#1:#3) arc (#1:#2:#3)}
	}
}

\ifshellescape
\usetikzlibrary{external}
\else
\PackageWarning{}{Shell escape is not enabled. Externalization won't work.}
\fi

\definecolor{colorinnertube}{RGB}{0,72,119}
\definecolor{coloroutertube}{RGB}{200,34,84}
\definecolor{udsblue}{RGB}{0,72,119}
\definecolor{udsred}{RGB}{200,34,84}
\definecolor{tuigreen}{RGB}{0,139,138}
\definecolor{udsyellow}{RGB}{215,223,35}

\newcommand{\figref}[1]{Figure~\ref{#1}}

\DeclareMathOperator*{\argmin}{arg\, min}

\definecolor{matlabyellow}{RGB}{237,177,32}

\newcommand{\mbold}[1]{\boldsymbol{#1}}

\newcommand{\p}{{\rho}}

\newcommand{\SO}{\mathrm{SO(3)}}

\newcommand{\Rm}{{R}}
\newcommand{\bolde}{\mbold{e}}
\newcommand{\boldu}{{u}}

\definecolor{udsblue}{RGB}{0,72,119}
\definecolor{udsred}{RGB}{200,34,84}
\definecolor{tuigreen}{RGB}{0,139,138}
\definecolor{udsgreen}{RGB}{222,158,10}
\definecolor{udsorange}{RGB}{0,76,0}

\newcommand{\R}{\ensuremath{\mathbb{R}}\xspace}

\newcommand{\norm}[2][2]{\left\|#2\right\|_#1}
\newcommand{\normexp}[2][2]{\left\|#2\right\|_#1^#1}
\newcommand{\sbi}{\ensuremath{s^\mathrm{b}_i}\xspace}
\newcommand{\bgi}{\ensuremath{\beta_{\gamma,i}}\xspace}

\newcommand{\dgi}{\ensuremath{d_{\gamma}}\xspace}
\newcommand{\jgi}[1][i]{\ensuremath{j_{\gamma,#1}}\xspace}
\newcommand{\jgis}[1][i]{\ensuremath{\jgi[#1]^\star}\xspace}

\newcommand{\setI}{\ensuremath{\mathbb{I}_{\gamma}}\xspace}
\newcommand{\setJ}{\ensuremath{\mathbb{J}}\xspace}

\newcommand{\setG}{\ensuremath{\mathbb{G}}\xspace}

\newcommand{\srj}{s^\mathrm{r}_j}

\newcommand{\bjstar}{\bj^\star}
\newcommand{\OP}{\ensuremath{\mathrm{(OP)}}\xspace}

\newcommand{\os}{One-Step\xspace}
\newcommand{\ms}{Multi-Step\xspace}

\newcommand{\bj}{\ensuremath{\boldsymbol{j}}\xspace}

\newcommand{\githublink}{\url{https://github.com/MKHoffmann/icpReconstructor}}

\newcommand{\mcode}[1]{\lstinline[style=Matlab-editor]!#1!}

\newtheorem{method}{Algorithm}[]
\usepackage[colorinlistoftodos,textsize=tiny]{todonotes} 
\usepackage{ifthen} 

\newboolean{correctionsApproved} 
\setboolean{correctionsApproved}{true} 
\newboolean{remarksApproved} 
\setboolean{remarksApproved}{true} 

\newcommand{\colorKFcorrection}{blue} 
\newcommand{\colorKFremark}{blue!50} 
\newcommand{\colorMKHcorrection}{red} 
\newcommand{\colorMKHremark}{red!50} 
\newcommand{\colorJMcorrection}{green!50!black} 
\newcommand{\colorJMremark}{green!25!black} 

\newcommand{\colorDZHcorrection}{orange} 

\ifthenelse{\boolean{correctionsApproved} \and \boolean{remarksApproved}} 
{}
{} 

\newcommand{\KFcorrection}[2][]{%
	\ifthenelse{\boolean{correctionsApproved}}%
	{\ignorespaces#2\ignorespaces\xspace}
	{%
		{\color{\colorKFcorrection}{#2}}
		\addcontentsline{tdo}{todo}{\color{\colorKFcorrection}{KF correction:} #1\xspace}
	}%
}

\newcommand{\KFremark}[2][]{%
	\ifthenelse{\boolean{remarksApproved}}%
	{~\ignorespaces}
	{%
		\footnote{\color{\colorKFremark}{#2}}
		\addcontentsline{tdo}{todo}{\color{\colorKFremark}{KF remark:} #1}
	}%
}

\newcommand{\MKHcorrection}[2][]{%
	\ifthenelse{\boolean{correctionsApproved}}%
	{\ignorespaces#2\ignorespaces\xspace}
	{%
		{\color{\colorMKHcorrection}{#2}}
		\addcontentsline{tdo}{todo}{\color{red}{MKH correction:} #1\xspace}
	}%
}

\newcommand{\MKHremark}[2][]{%
	\ifthenelse{\boolean{remarksApproved}}%
	{~\ignorespaces}
	{%
		\footnote{\color{\colorMKHremark}{#2}}
		\addcontentsline{tdo}{todo}{\color{\colorMKHremark}{MKH remark:} #1}
	}%
}

\newcommand{\JMcorrection}[2][]{%
	\ifthenelse{\boolean{correctionsApproved}}%
	{\ignorespaces#2\ignorespaces\xspace}
	{%
		{\color{\colorJMcorrection}{#2}}
		\addcontentsline{tdo}{todo}{\color{red}{MKH correction:} #1\xspace}
	}%
}

\newcommand{\JMremark}[2][]{%
	\ifthenelse{\boolean{remarksApproved}}%
	{~\ignorespaces}
	{%
		\footnote{\color{\colorJMremark}{#2}}
		\addcontentsline{tdo}{todo}{\color{\colorJMremark}{MKH remark:} #1}
	}%
}

\newcommand{\DZHcorrection}[2][]{%
    \ifthenelse{\boolean{correctionsApproved}}%
	{\ignorespaces#2\ignorespaces\xspace}
	{%
		{\color{\colorDZHcorrection}{#2}}
		\addcontentsline{tdo}{todo}{\color{red}{MKH correction:} #1\xspace}
	}%
}

\let\oldlistoftodos\listoftodos
\renewcommand{\listoftodos}{%
	\ifthenelse{\boolean{correctionsApproved} \and \boolean{remarksApproved}}%
	{}
	{\oldlistoftodos}
}

\usepackage{float}
\usepackage{tabularx}

\usepackage[backend=biber,style=numeric-comp,sorting=none,maxbibnames=99]{biblatex}

\newcommand{\journaltitle}{Mathematical and Computer Modelling of Dynamical Systems}
\newcommand{\articledoi}{10.1080/13873954.2026.2716623}
\edef\doiurl{https://doi.org/\articledoi}
\newcommand{\pubdate}{17 August 2026}

\theoremstyle{plain}
\newtheorem{theorem}{Theorem}[section]

\theoremstyle{definition}

\theoremstyle{remark}

\begin{document}

\title{Reconstruction of continuum robots by marker-free shape registration of image data using a kinematic model}

\author{
	Matthias K. Hoffmann\thanks{Corresponding author email: matthias.hoffmann@uni-saarland.de}\\
	Systems Modeling and Simulation\\
	Saarland University\\
	Saarbr\"ucken, Germany
	\And
	Julian M\"uhlenhoff\\
	Mechatronics Group\\
	Technische Universit\"at Ilmenau\\
	Ilmenau, Germany
	\And
	Zhaoheng Ding\\
	Mechatronics Group\\
	Technische Universit\"at Ilmenau\\
	Ilmenau, Germany
	\AND
	Thomas Sattel\\
	Mechatronics Group\\
	Technische Universit\"at Ilmenau\\
	Ilmenau, Germany
	\And
	Kathrin Fla\ss kamp\\
	Systems Modeling and Simulation\\
	Saarland University\\
	Saarbr\"ucken, Germany
}

\graphicspath{{figs}}


\begin{acronym}
	\acro{icp}[ICP]{Iterative Closest Point}
\end{acronym}

\maketitle

\begin{abstract}
Continuum robots are slender, flexible manipulators that navigate confined, curved workspaces and are gaining traction in aerospace, inspection, automation, and minimally invasive medical applications. Predicting their shape from physics-based models alone remains challenging, making accurate measurement of the deformed backbone essential for model validation and reference-data acquisition. We present an optimization-based shape-registration algorithm that fits a parametric three-dimensional curve directly to image observations within a photogrammetric pipeline, targeting marker-free measurement rather than sensing under occlusion. By matching reconstruction points to robot pixels, the method requires no prior knowledge of the robot's location in each image. Across most configurations, the estimated backbone deviates from ground truth by less than \SI{1}{\milli\meter} (\SI{0.67}{\percent} of the robot's length). On real concentric-tube continuum robots, the reconstruction agrees with ten discrete manual photogrammetric measurements over 18 configurations, while replacing the manual procedure with an automated pipeline that runs in roughly \SI{0.5}{\second} per configuration.
\end{abstract}


\begin{center}
	\fbox{\parbox{0.95\textwidth}{\small
			This is an original manuscript of an article published by Taylor \& Francis in \emph{\journaltitle} on \pubdate, available online: \expandafter\url\expandafter{\doiurl}.
	}}
\end{center}


\section{Introduction}\label{sec_Intro}
\noindent
Continuum robots (CRs) have received increasing attention for applications in aerospace, medical surgery, automation, and inspection \cite{russo2023continuumrobots}. Inspired by biomechanical structures such as snakes, elephant trunks, and tentacles, they replace conventional joints with a continuously deformable centre line, the \emph{backbone}, whose trajectory governs the robot's motion and therefore its applicability \cite{BurgnerKahrs2014}. CR variants differ mainly in actuation principle \cite{russo2023continuumrobots} but share a slender, continuously differentiable outer shape.

Accurate knowledge of tip and backbone position is required to reach targets and maneuver around obstacles \cite{BurgnerKahrs2014}. This drives extensive work on CR modeling \cite{webster2010design,barrientos2021model,rucker2010basemodel,lamping2022pneumatic,Watts2019,GreinerPetter2017} and, equally important, on backbone measurement for model validation and closed-loop control.

\subsection{Shape estimation of continuum robots}\label{sec_shape_estimation}
\noindent
In literature, there is a wide variety of measurement principles applied to CRs \cite{Sincak2024}. Especially for surgical applications, medical imaging devices such as optical coherence tomography \cite{Baran2017}, X-ray based imaging \cite{Jayender2008,Keiner2025}, magnetic resonance imaging \cite{Battisti2016,Su2012,burgner2011mri} and ultrasound \cite{Abayazid2013,Nadeau2015} are used.
These imaging devices are not hindered by surrounding tissue and support a possible application as a medical product \cite{Sorriento2020}, however, resolution is usually limited to roughly $1\,\mathrm{mm}$ and, depending on the interaction between deployed materials and medical imaging type, artifacts in the image are likely to occur \cite{Keiner2025}.


For closed-loop deployment under obstructed-view conditions, several internal sensing types have been suggested, such as fiber Bragg grating \cite{Battisti2016,Ryu2014,Xu2016}, opto-electronic foams \cite{Meerbeek2018}, capacitive \cite{hu2023stretchable} and magnetic systems \cite{Adamu2025}, and others \cite{Wang2018}. They lack the applicability to slender CRs \cite{Meerbeek2018,hu2023stretchable} and are limited in backbone resolution \cite{Adamu2025,Meerbeek2018} and overall accuracy \cite{Adamu2025,Wang2018,hu2023stretchable}. Other measurement systems specifically designed for a certain type of CR, e.g.~for tendon-actuated continuum robots \cite{Rone2013,Feng2025}, are not generically enough to be applied to general CRs.
Electromagnetic tracking systems hinder the robot from moving freely \cite{Grassmann2018,Grassmann2022} and also yield only a small number of discrete backbone samples \cite{BurgnerKahrs2014,Grassmann2018,Grassmann2022,Xu2013} and therefore cannot reproduce the full backbone shape.
Moreover, all onboard sensing modalities require physically integrating sensors into the robot, which alters its mechanical behaviour and introduces forces that are difficult to capture in standard CR models such as \cite{rucker2010basemodel}.

Camera-based approaches overcome these limitations and are generally credited with higher achievable accuracy \cite{Glossop2009}, provided the CR is fully visible; additional cameras can compensate for partial occlusion.
Photogrammetric errors with medium- to high-quality optics lie in the range of 10 to 100~\si{\micro\meter} \cite{GreinerPetter2017,Muehlenhoff2022,Muehlenhoff2024,Luhmann2016}, sufficient for typical CR applications \cite{russo2023continuumrobots,BurgnerKahrs2014,Keiner2025}. When the CR is surrounded by translucent media, light-diffraction correction has been proposed but introduces additional error sources \cite{Watts2019,Kallem2009}. 
Even when deployment requires insertion into opaque media where vision-based approaches cannot be applied, they are a corner stone for establishing and validating highly accurate models \cite{Abayazid2013,Chitrakaran2004,Webster2009,rucker2010basemodel,Gilbert2015,GreinerPetter2017,Watts2019,Kuntz2020,Muehlenhoff2022}, which in turn allow more precise movements of CRs.

The central difficulty of camera-based reconstruction is identifying corresponding 2D image points across cameras so that the underlying 3D point can be triangulated.
In some existing pipelines, this step is performed manually: a human operator picks matching pixels in each image, typically supported by tools that constrain the search rather than perform it automatically -- pre-attached reference markers act as visual anchors \cite{Webster2009, rucker2010basemodel}, an epipolar-line search narrows the candidate region to a single line in the second image \cite{Gilbert2015}, or only the robot tip is annotated, which is unique whenever it is visible in all views \cite{Abayazid2013,Chitrakaran2004}. Other works automate the selection by attaching physical markers along the backbone -- polystyrene balls \cite{GreinerPetter2017,Muehlenhoff2022,Su2012,Xu2008}, chessboards \cite{Muehlenhoff2022}, paint \cite{rucker2010basemodel}, or LEDs \cite{pedari2019spatial} -- that are easy to automatically detect and match.
Both routes are problematic: manual annotation is laborious and introduces human error, while attached markers alter the robot's behaviour, prevent certain configurations such as full tube retraction in concentric tube continuum robots, and may occlude one another in the camera views \cite{GreinerPetter2017, Muehlenhoff2022}.

Marker-free algorithms avoid these drawbacks by fitting models to image observations. 
Hannan et al.\ \cite{hannan2005real} chain circular arcs to the separating disks of a trunk-type robot in 2D.
Camarillo et al.\ \cite{camarillo2008vision} and Kuntz et al.\ \cite{Kuntz2020} both use shape-from-silhouette: voxels of a 3D grid are projected onto the images, those matching the robot are kept, and a space curve is fit through the resulting volume (via moving average and ordinary least squares, respectively). 
Lu et al.\ \cite{lu2023diffrendering} parameterize the backbone with Bézier curves inside a differentiable renderer and minimize the per-pixel difference between rendered and observed images, enabling gradient-based optimization of the curve parameters.

\renewcommand{\arraystretch}{1.}
\begin{table}[htbp]
	\small
	\centering
	\caption{Comparison with existing reconstruction methods for continuum robots}
	\label{tab:reconstr_methods}
	\begin{tabular}{|l | c | c | c | c | c | c | c | c | c |}
		\hline {Source} & \cite{Webster2009, rucker2010basemodel} & \cite{Gilbert2015} & \cite{Abayazid2013} & \cite{Chitrakaran2004} & \cite{GreinerPetter2017} & \cite{hannan2005real} & \cite{camarillo2008vision} & \cite{lu2023diffrendering} & Ours \\
		\hline {1. Algorithm is fully automated} & \xmark & \xmark & \cmark & ? & \cmark & \cmark & \cmark & \cmark & \cmark\\
		\hline {2. Algorithm does not require markers} & \xmark & \xmark & \cmark & ? & \xmark & \cmark & \cmark & \cmark & \cmark\\
		\hline {3. Minimum number of cameras} & 2 & 2 & 1 & 1 & 2 & 1 & 3 & 1 & 2\\
		\hline {4. Gives results in three dimensions} & \cmark & \cmark & \xmark & \cmark & \cmark & \xmark & \cmark & \cmark & \cmark\\
		\hline {5. Works without prior knowledge} & \cmark & \cmark & \cmark & \cmark & \cmark & \cmark & \cmark & \xmark & \cmark \\
		\hline {6. Allows for incorporating robot specific features} & \xmark & \xmark & \xmark & \xmark & \xmark & \xmark &\xmark & \cmark & \cmark \\
		\hline {7. Gives a description of the full backbone} & \cmark & \cmark & \xmark & \cmark & \xmark & \cmark & \cmark & \cmark & \cmark\\
		\hline {8. Algorithm code is published} & \xmark & \xmark & \xmark & \xmark & \xmark & \xmark & \xmark & \xmark & \cmark\\
		\hline
	\end{tabular}
\end{table}


\subsection{Contribution}\label{sec_contribution}
\noindent
The literature lacks a robust, automated, marker-free method that recovers the full three-dimensional backbone of a CR from camera views with low measurement error.
We consider the following setting: the robot is observed by two or more calibrated cameras with an essentially unobstructed view, it carries neither markers nor onboard shape sensors, and the quantity of interest is the geometric shape of the backbone itself rather than the underlying robot state. 
A representative use case is the validation of kinematic or mechanical CR models, where an independent, accurate measurement of the deformed shape is needed for comparison.
We present an optimization-based algorithm that closes this gap: a parametric 3D space curve is fit directly to binary images of the robot and projected onto the image planes, with the curve generated by the moving-frame formulas.
These differential equations -- not previously used for shape reconstruction, but highly prevalent in CR modeling -- produce smooth, equidistantly samplable curves and align more closely with established CR kinematic models than polynomial parameterizations do.

In contrast to differentiable rendering \cite{lu2023diffrendering}, we operate directly on the backbone and avoid constructing a full volumetric body, reducing computational cost.
The algorithm is validated on both artificial and real images of concentric tube continuum robots (CTCRs) and the source code is released open-source\footnote{\githublink} to enable reuse and direct comparison.
Table~\ref{tab:reconstr_methods} positions the proposed method against the prior work discussed above.
Along eight criteria, ranging from automation and marker requirements to publication of code; question marks denote criteria for which the cited source provides no information.

\subsection{Outline}
\noindent
Section~\ref{sec:optimal} states the reconstruction problem in a general setting, motivates the moving-frame backbone model and the image projection, and proves convergence of the algorithm.
Section~\ref{sec:results_optimal} reports numerical results on artificial CTCR images; Section~\ref{sec:validation} validates the method on real CTCR images; and Section~\ref{sec:summary_outlook} summarizes the findings and outlines future work.


\section{Optimization-based 3D pose estimation}
\label{sec:optimal}
\noindent
We propose an optimization-based reconstruction algorithm that, similar to the differentiable rendering approach of Lu et al.\,~\cite{lu2023diffrendering}, fits a parametric 3D curve to the image observations but, unlike differentiable rendering, does not require generating an intermediate 3D volume.
Concretely, we project the parametric curve onto the image planes and match its projections to the observed image content, minimizing an error measure in a least-squares sense.

A complication of working directly in image space is that the projection from 3D to the image planes is many-to-one: the arc length of the robot corresponding to a given pixel is unknown.
We address this with an iterative closest point (ICP) formulation that pairs each robot pixel with the closest projected reconstruction point and refines the pairing as the curve evolves.

\subsection{Problem formulation} \label{sec:Probform}

We begin by stating the reconstruction problem in a general setting, before deriving the ICP-based scheme that handles the unknown pixel-to-arc-length correspondence.
Let two curves be parametrized by their respective arc lengths \( s \): the first, \( b \), represents the actual backbone of the robot, and the second, $r$, is the one-dimensional reconstruction we aim to determine by optimizing the $m$ decision variables \( \theta \) that parameterize its shape.\ These curves are defined as 
\(
	b:\ [0,L] \rightarrow \R^3, \
	s\mapsto b(s),
\)
and
\(
	r:\ [0,L]\times\R^m \rightarrow \R^3, \
	(s,\theta) \mapsto r(s;\theta),
\)
respectively, where $L$ is the known length of the backbone.
Solving the parameter identification problem for \(r(\cdot;\theta)\), the difference between the actual backbone and the reconstructed curve has to be minimized over the length of the entire backbone. 
Considering any $p \ge 1$ for the considered $p$-norm, the unconstrained optimization problem reads as
\begin{align*}
	\min_\theta \ & \int_{0}^{L} \normexp[p]{ b(s) - r(s; \theta)} \, ds .
\end{align*}
However, several challenges arise when applying this formulation to {shape registration of} continuum robots:
\begin{enumerate}
	\item We observe the backbone \( b(s) \) {only} through \( C \) known projections \( T_\gamma(b(s)) \), with the projection function \( T_\gamma: \R^3 \rightarrow \R^2 \) mapping 3D points to their 2D image representations, where \( \gamma \in \{1, \ldots, C\} \).
	\item The projections \( T_\gamma \) map to images, i.e.\ a finite number of discrete pixels.
Thus, the \( M_\gamma \) observed pixels that correspond to the robot are denoted as \( T_\gamma(b(\sbi)) \), where $\sbi \in [0, L] \, \forall \, i\in \left\lbrace1, 2, \dots, M_\gamma\right\rbrace$ are unknown arc length values.
	\item As the values of \( \sbi \) are unknown, it is impossible to match the projections of \( b \) and \( r \) directly at corresponding arc lengths.
\end{enumerate}
Let us define the following index sets
\begin{align*}
	\setG &= \left\lbrace 1, 2, ..., C \right\rbrace,\\
	\setI &=  \left\lbrace 1, 2, ..., M_\gamma \right\rbrace {\text{ for } \gamma \in \setG },\\ 
	\setJ &= \left\lbrace 1, 2, ..., N \right\rbrace ,
\end{align*}
containing the indices of the $C$ images, the indices of the $M_\gamma$ pixels corresponding to the robot per image, and the indices of the $N$ discrete fixed-arc-length reconstruction points, respectively.
This gives the sets of robot and reconstruction image coordinates 
\begin{align*}
	B_\gamma &= \left\lbrace \bgi = T_\gamma(b(\sbi)) \, \left| \, i \, \in \setI \right. \right\rbrace, \\ 
	R_\gamma(\theta) &= \left\lbrace \nu_{\gamma,j} = T_\gamma(r(\srj; \theta)) \, \left| \, j \, \in \, \setJ \right. \right\rbrace,
\end{align*}
where $\left\lbrace s^\mathrm{r}_1, s^\mathrm{r}_2, \dots, s^\mathrm{r}_N\right\rbrace$ is the set of $N$ fixed arc lengths at which $r$ is evaluated.

\subsection*{Iteratively measuring the model mismatch}
To resolve the unknown correspondences listed above, we base the correspondence search on the ICP approach of Besl and McKay \cite{besl1992icp}, originally formulated for rigid alignment of two point clouds: for each point of the reference cloud $R$ the closest point in the data cloud $B$ is found, and the rigid transformation that minimizes the sum of these \(R\)-to-\(B\) distances is computed.
Subsequent work \cite{godin1994invcorrespondence} showed that additionally including the reverse, \(B\)-to-\(R\) correspondence reduces local minima and aids convergence; for an overview of rigid and non-rigid variants see \cite{icpreview2013tam}. We adopt the \(B\)-to-\(R\)-formulation and match each backbone pixel \( T_\gamma(b(\sbi)), \, i \in \mathbb{I}_\gamma \) to the closest point in the discretized and projected reconstruction \( T_\gamma(r(\srj)), \, j \in \mathbb{J}\); the released implementation, however, supports rapid prototyping of alternative formulations.
Although initial pairings of points are generally incorrect, iterative refinement converges to a consistent solution, illustrated in Figure~\ref{fig:reconstruction_optimal}.

\begin{figure*}
	\centering


\begin{tikzpicture}[scale=0.9,
	arr/.style={
		-{Latex[length=4mm, width=2mm]}, 
		line width=2pt,       
		color=black           
	}]
\newcommand{\rrval}{7}
\begin{axis}[clip mode=individual,
	axis x line=center,
	axis y line=center,
	axis equal,
	xticklabels={,,},
	yticklabels={,,},
	axis line style={draw=none},
	ticks=none,
	legend columns=3,
	legend style={anchor=south,legend cell align=left, cells={align=left}, column sep=0.94ex},
	legend entries={robot, robot pixels, robot estimate, estimate pixels, estimate error per pixel},
	at={(200, 150)}, xshift=-5mm]
	\addlegendimage{udsred, opacity=0.2, line width=4}
	\addlegendimage{only marks, mark=*, mark size=2pt, udsred}
	\addlegendimage{udsblue, opacity=0.2, line width=4}
	\addlegendimage{only marks, mark=*, mark size=2pt, udsblue}
	\addlegendimage{arr}
	\addplot[draw=none] {x};
\end{axis};
\begin{groupplot}[
	group style={group size=3 by 1,
				horizontal sep = -1.5cm,
				},
	xmin=4,
	xmax=6,
	ymin=-11,
	ymax=0,
	clip mode=individual,
	axis x line=center,
	axis y line=center,
	axis equal,
	xticklabels={,,},
	yticklabels={,,},
	axis line style={draw=none},
	ticks=none,
	width=8cm
	]
\nextgroupplot[title = Initial guess, title style={anchor=south east}]

\pgfmathsetmacro{\rb}{50}
\pgfmathsetmacro{\rr}{\rrval}
\pgfmathsetmacro{\angend}{360-(\rr/\rb)*180/2}
\pgfmathsetmacro{\nb}{7}
\pgfmathsetmacro{\nr}{7}
\pgfmathsetmacro{\db}{(360-\angend)/(\nb-1)}
\pgfmathsetmacro{\dr}{90/(\nr-1)}

\addplot[domain=360:270, samples=100, udsred, opacity=0.2, line width=4] ({\rr*cos(x)},{\rr*sin(x)});

\addplot[domain=360:270, samples=\nr, only marks, mark=*, mark size=2pt, udsred] ({\rr*cos(x)},{\rr*sin(x)});

\addplot[domain=360:\angend, samples=100, udsblue, opacity=0.2, line width=4] ({\rb*cos(x)-(\rb-\rr)},{\rb*sin(x)});
\addplot[domain=360:\angend, samples=\nb, only marks, mark=*, mark size=2pt, udsblue] ({\rb*cos(x)-(\rb-\rr)},{\rb*sin(x)});

\node[] (tipr) at (axis cs: {\rr*cos(270)},{\rr*sin(270)}) {};
\node[] (pretipb) at (axis cs: {\rb*cos(\angend+\db)-(\rb-\rr)},{\rb*sin(\angend+\db)}) {};
\draw[arr] ($(pretipb.center)!0.94!(tipr)$) -- ($(pretipb.center)!0.06!(tipr)$);

\pgfplotsinvokeforeach{1,...,4}
{
	\node[] (r#1) at (axis cs: {\rr*cos(360-\dr*(#1))},{\rr*sin(360-\dr*(#1))}) {};
	\node[] (b#1) at (axis cs: {\rb*cos(360-\db*(#1))-(\rb-\rr)},{\rb*sin(360-\db*(#1))}) {};
	\draw[arr] ($(r#1.center)!0.09!(b#1.center)$) -- ($(b#1.center)!0.06!(r#1.center)$);
}

\pgfplotsinvokeforeach{5,...,\nb-1}
{
	\node[] (r#1) at (axis cs: {\rr*cos(360-\dr*(#1))},{\rr*sin(360-\dr*(#1))}) {};
	\node[] (b#1) at (axis cs: {\rb*cos(360-\db*(#1-1))-(\rb-\rr)},{\rb*sin(360-\db*(#1-1))}) {};
	\draw[arr] ($(r#1.center)!0.09!(b#1.center)$) -- ($(b#1.center)!0.06!(r#1.center)$);
}

.\nextgroupplot[title = Step 1, title style={anchor=south east}]

\pgfmathsetmacro{\rb}{10}
\pgfmathsetmacro{\rr}{\rrval}
\pgfmathsetmacro{\angend}{360-(\rr/\rb)*180/2}
\pgfmathsetmacro{\nb}{7}
\pgfmathsetmacro{\nr}{7}
\pgfmathsetmacro{\db}{(360-\angend)/(\nb-1)}
\pgfmathsetmacro{\dr}{90/(\nr-1)}

\addplot[domain=360:270, samples=100, udsred, opacity=0.2, line width=4] ({\rr*cos(x)},{\rr*sin(x)});

\addplot[domain=360:270, samples=\nr, only marks, mark=*, mark size=2pt, udsred] ({\rr*cos(x)},{\rr*sin(x)});

\addplot[domain=360:\angend, samples=100, udsblue, opacity=0.2, line width=4] ({\rb*cos(x)-(\rb-\rr)},{\rb*sin(x)});
\addplot[domain=360:\angend, samples=\nb, only marks, mark=*, mark size=2pt, udsblue] ({\rb*cos(x)-(\rb-\rr)},{\rb*sin(x)});

\pgfplotsinvokeforeach{1,...,\nb-1}
{
	\node[] (r#1) at (axis cs: {\rr*cos(360-\dr*#1)},{\rr*sin(360-\dr*#1)}) {};
	\node[] (b#1) at (axis cs: {\rb*cos(360-\db*#1)-(\rb-\rr)},{\rb*sin(360-\db*#1)}) {};
	\draw[arr] ($(r#1.center)!0.09!(b#1.center)$) -- ($(b#1.center)!0.06!(r#1.center)$);      
}

\nextgroupplot[title = Step 2, title style={anchor=south east}]
	\pgfmathsetmacro{\rb}{8}
	\pgfmathsetmacro{\rr}{\rrval}
	\pgfmathsetmacro{\angend}{360-(\rr/\rb)*180/2}
	\pgfmathsetmacro{\nb}{7}
	\pgfmathsetmacro{\nr}{7}
	\pgfmathsetmacro{\db}{(360-\angend)/(\nb-1)}
	\pgfmathsetmacro{\dr}{90/(\nr-1)}
	
	\addplot[domain=360:270, samples=100, udsred, opacity=0.2, line width=4] ({\rr*cos(x)},{\rr*sin(x)});
	
	\addplot[domain=360:270, samples=\nr, only marks, mark=*, mark size=2pt, udsred] ({\rr*cos(x)},{\rr*sin(x)});
	
	\addplot[domain=360:\angend, samples=100, udsblue, opacity=0.2, line width=4] ({\rb*cos(x)-(\rb-\rr)},{\rb*sin(x)});
	\addplot[domain=360:\angend, samples=\nb, only marks, mark=*, mark size=2pt, udsblue] ({\rb*cos(x)-(\rb-\rr)},{\rb*sin(x)});
	
	\addlegendimage{arr}
	
	\pgfplotsinvokeforeach{1,...,\nb-1}
	{
		\node[] (r#1) at (axis cs: {\rr*cos(360-\dr*#1)},{\rr*sin(360-\dr*#1)}) {};
		\node[] (b#1) at (axis cs: {\rb*cos(360-\db*#1)-(\rb-\rr)},{\rb*sin(360-\db*#1)}) {};
		\draw[arr] ($(r#1.center)!0.09!(b#1.center)$) -- ($(b#1.center)!0.06!(r#1.center)$);      
	}
	
\end{groupplot}

\end{tikzpicture}
	\caption{Matching two circular arcs of the same length with $\theta$ as the radius.
The cost function is computed as the sum of all reconstruction errors using the \( B \)-to-\( R \)-correspondence.\
	The reconstruction error is the distance of each cannula pixel to the closest reconstruction pixel.\ 
	Initial matching errors like in the initial guess decrease over the iteration steps, so that in the end, in this example, the estimate is ideal.}
	\label{fig:reconstruction_optimal}
\end{figure*}

For the \( i \)-th pixel's coordinates \bgi in image $\gamma$, the distance measure to the \( j \)-th reconstruction point is given by
\begin{align}
	\dgi(\theta, j, i) = \left\|\bgi - \nu_{\gamma,j}\right\|_p^p,\ j \in \setJ, \ \forall \ i\in \setI. \label{eq:dist}
\end{align}

The cost function to minimize {for \(r(\cdot;\theta) \) identification} is
\begin{align*}
	J(\theta, \bj_1, \dots, \bj_C) = \sum_{\gamma\in\setG}\ \sum_{i\in\setI} \dgi(\theta, \jgi, i),
\end{align*}
where
\[ \bj_\gamma = \left\lbrace \jgi[1], \jgi[2], \ldots, \jgi[M_\gamma] \ | \ \jgi \in \setJ \ \forall \ i \in \setI \right\rbrace \]
is a set of indices that assigns to each pixel in $B_\gamma$ a point from $R_\gamma(\theta)$.
In fact, we are interested in assigning the \emph{closest} point from $R_\gamma(\theta)$ to each pixel in $B_\gamma$.
Incorporating this into the optimization leads to the bi-level problem
\begin{align*}
	\min_\theta \ & J(\theta, \bjstar_1, \dots, \bjstar_C)\\
	\text{s.t.} \ & \jgi^\star  = \argmin_{j\in\setJ} \dgi(\theta, j, i), \ \forall \ i\in \setI, \ \forall \ \gamma \in \setG.
\end{align*}
This problem is challenging due to the non-differentiability of the optimization problem in the constraints.
To simplify, we replace \( \theta \) with a current estimate \( \tilde{\theta} \) in the constraint, yielding the ICP curve fitting
\begin{subequations}
\begin{align}
	\OP\qquad\ \nonumber\\
	\min_\theta \ & J(\theta, \bjstar_1, \dots, \bjstar_C) \label{eq:cost_j_approx}\\
	\text{s.t.} \ & \jgi^\star  = \argmin_{j\in\setJ} \dgi(\tilde\theta, j, i), \ \forall \ i\in \setI, \ \forall \ \gamma \in \setG. \label{eq:const_j_approx}
\end{align}
\end{subequations}
Since \eqref{eq:const_j_approx} no longer depends on $\theta$, (OP) becomes a standard curve-fitting problem with (most likely incorrectly) matched data.
After convergence, $\tilde{\theta}$ is reinitialized with the solution and the lower-level problem is solved again, yielding the recursive sequence in Algorithm~\ref{met:ms}, which we refer to as the \textit{\ms} formulation.
\begin{method}[\ms ICP curve-fitting]
\begin{subequations}
	\begin{align}
	\tilde{\theta}\left[k+1\right] & = \argmin_\theta \ J(\theta, \bjstar_1[k], \dots, \bjstar_C[k]) \label{eq:opt_simple}\\
	\jgi^\star[k] & = \argmin_{j\in\setJ}  \dgi(\tilde{\theta}[k], j, i), \ \forall \ i\in \setI, \ \forall \ \gamma \in \setG \label{eq:opt_simple_constr}\\
	\tilde{\theta}\left[0\right] & = \tilde{\theta}_0 \label{eq:opt_simple_init}
\end{align}
\end{subequations}
\label{met:ms}
\end{method}
\vspace{-6mm}
Besl and McKay \cite{besl1992icp} proved convergence of ICP for rigid least-squares registration: the curve-fitting step reduces the mean squared error between the two point clouds, while the closest-point step reduces each individual distance and therefore also the mean squared error.
The argument extends to non-linear curves $r(s;\theta)$: provided the inner solver does not increase $J$, the iterate $\theta[k]$ is a local minimizer at fixed $\bjstar_\gamma[k]$. In subsequent iterations $\bjstar_\gamma$ is recomputed by exhaustive comparison of all projected reconstruction points with their counterparts in $B_\gamma$; since the $\dgi(\tilde\theta[k],\jgis,i)$ are independent in $\jgis$, reassignment changes only individual $\dgi$ terms, hence the recursive sequence converges.

\begin{theorem}[Convergence of the ICP curve-fitting]
	The ICP curve-fitting approach for our problem setting \eqref{eq:opt_simple}-\eqref{eq:opt_simple_init} converges under the same conditions as those stipulated for solving the curve-fitting problem \eqref{eq:opt_simple}.
\end{theorem}
\begin{proof}
	The cost function \( J(\tilde\theta[k], \bjstar_1[k], \dots, \bjstar_C[k]) \) is the sum of distances\\ \( \dgi(\tilde\theta[k], \jgis[k], i) \). These terms are independent of each other with respect to their corresponding $\jgis[k]$. Minimizing each \( \dgi(\tilde\theta[k], j, i)) \) during \eqref{eq:opt_simple_constr} with respect to $j$ leads to the minimizer index \( \jgi^\star[k] \). If \( {\jgis}[k-1] = {\jgis}[k] \), then \( d_i(\tilde\theta[k], {\jgis}[k-1], i) = d_i(\tilde\theta[k], {\jgis}[k], i) \). Conversely, if \( {\jgis}[k-1] \neq {\jgis}[k] \), then \( d_i(\tilde\theta[k], {\jgis}[k], i) \leq d_i(\tilde\theta[k], {\jgis}[k-1], i) \), as otherwise \( {\jgis}[k] \) would not be a minimizer of equation~\eqref{eq:opt_simple_constr}.\ 
	Therefore, \( \sum_{i=1}^M d_i(\tilde\theta[k], \bjstar_1[k], \dots, \bjstar_C[k]) \leq J(\tilde\theta[k], \bjstar_1[k-1], \dots, \bjstar_C[k-1]) \), ensuring that updating the correspondences in the described manner does not increase the overall cost function.\ 
	Since $J$ is bounded from below with 0 and decreases with the reassignment of correspondences, the convergence of the algorithm is guaranteed, provided that the iteration step \eqref{eq:opt_simple} does not increase the cost function.
\end{proof}
\vspace{-2.5mm}
The same argument shows that any iterative scheme that does not increase the cost in \eqref{eq:opt_simple} converges.
In particular, solving \eqref{eq:opt_simple} to full convergence between correspondence updates can miss the iterate at which a different reconstruction point becomes the closest.
We therefore also consider a simpler scheme that takes a single gradient-descent step per iteration, yielding the \textit{\os} formulation in Algorithm~\ref{met:os}.
\begin{method}[\os ICP curve-fitting]
\begin{align*}
	\tilde{\theta}[k+1] & = \tilde{\theta}[k] - \alpha \left. \frac{\partial J(\theta, \bjstar_1[k], \dots, \bjstar_C[k])}{\partial \theta} \right|_{\theta = \tilde{\theta}[k]} \\
	\jgi^\star[k] & = \argmin_{j\in\setJ}  \dgi(\tilde{\theta}[k], j, i), \ \forall \ i\in \setI, \ \forall \ \gamma \in \setG \\
	\tilde{\theta}[0] & = \tilde{\theta}_0
\end{align*}
\label{met:os}
\end{method}
Applying {a typical} 
 Armijo condition to the step length\footnote{In the machine learning context often referred to as learning rate.} $\alpha$ 
  ensures sufficient decrease at each step, thereby guaranteeing convergence of this iterative process \cite{nocedal1999numerical}.

\subsection{A kinematic model for easy sampling}
\noindent
The development so far is independent of how $r$ is modeled.
Prior works use component-wise polynomials \cite{Kuntz2020, lu2023diffrendering}, but these do not admit equidistant sampling with respect to arc length, which is required to faithfully represent length measurements along the curve.
We instead adopt the kinematic model describing the moving-frame concept from differential geometry, which is also the basis of many physics-based CR models \cite{rucker2010basemodel, dupont2017kinematic, rao2021tendon, lamping2023novel, rucker2011tendon}: an orthonormal frame moves along its own $z$-axis while being rotated around its axes by the curvatures \( \boldu(s; \theta) = \begin{bmatrix} u_x(s; \theta) & u_y(s; \theta) & u_z(s; \theta) \end{bmatrix} \) (see \figref{fig:frame}).
Intuitively, integrating these formulas turns a curvature profile $\boldu(s;\theta)$ into a 3D backbone: the curvatures continuously reorient the moving frame, and following its local $z$-axis traces out the position $\p(s)$ along the arc length.

With $\theta$ again denoting the decision variables and $\p(s)$, $\Rm(s) = \begin{bmatrix}
	\bolde_x(s) & \bolde_y(s) & \bolde_z(s)
\end{bmatrix}$ the position and orientation of the frame at arc length $s$, we set $r(s;\theta) = \p(s)$, defined as the solution of
\begin{equation}
	\begin{aligned}
		r(s; \theta) &= \p(s),\\
		\p'(s) &= \bolde_z(s),\\
		\Rm'(s) &= \Rm(s)\hat{\boldu}(s; \theta),\\
		\p(0) &= \p^0 \in \R^3,\\
		\Rm(0) &= \Rm^0 \in \SO\\
		\text{with}\ \hat{\boldu}(s; \theta) &= \begin{bmatrix}
			0 & -u_z(s; \theta) & u_y(s; \theta) \\
			u_z(s; \theta) & 0 & -u_x(s; \theta) \\
			-u_y(s; \theta) & u_x(s; \theta) & 0
		\end{bmatrix}.
	\end{aligned}\label{eq:moving_frame}
\end{equation}

\begin{figure}
	\centering
	\input{figs/moving_frame_2.tex}
	\caption{A 3D curve $\rho (s)$ traced by a moving frame, expressed in the global coordinate system $(X,Y,Z)$.\ 
		The frame follows its local $z$-axis and rotates around the axes with the curvatures $u_x$, $u_y$, and $u_z$.}
	\label{fig:frame}
\end{figure}

The curvatures \( u_x \) and \( u_y \) are parameterized as piecewise functions of polynomials of degree {three}, 
for which $l[i]$ with $i \in \{1, 2, \dots, S\}$ denotes the length of the $S$ segments and \( l[0] = 0 \).\
The torsion \( u_z \) is not observable from the projected backbone shape and is therefore set to 0.
The parameters \( \theta \) for these polynomials are chosen to capture the curvature values and their derivatives at the arc lengths corresponding to the lengths of the previous and the current tube.
The parameter set \( \theta^i \) for the $i$-th piece's curvature polynomials is then defined as 
\begin{equation}
	\begin{aligned}
		\theta^i_j &= \begin{bmatrix}
			\theta^i_{j,1} & \theta^i_{j,2} & \theta^i_{j,3} & \theta^i_{j,4}
		\end{bmatrix} \\
		&= \begin{bmatrix}
			{u_j^i}(l[i-1]) & {u_j^i}'(l[i-1]) & {u_j^i}(l[i]) & {u_j^i}'(l[i])
		\end{bmatrix} \\
		&\quad \forall \ i \in \{1, 2, 3\}, \ j \in \{x, y\}.
	\end{aligned}
	\label{eq:params}
\end{equation}
Here, \( u_x(l[\cdot]) \) and \( u_y(l[\cdot]) \) represent the curvature values at the start and end of each piece, while \( u_x'(l[\cdot]) \) and \( u_y'(l[\cdot]) \) represent their respective derivatives at these points.\
The resulting polynomials are
\begin{align*}
	u_j^i(s; \theta) =\ &\frac{2\theta^i_{j,1} - 2\theta^i_{j,3} + \Delta l \theta^i_{j,2} + \Delta l \theta^i_{j,4}}{\Delta l^3}(s-l[i-1])^3 \\
	-\ &\frac{3\theta^i_{j,1} -3\theta^i_{j,3} + 2\Delta l\theta^i_{j,2} + \Delta l \theta^i_{j,4}}{\Delta l^2}(s-l[i-1])^2 \\
	+\ & \theta^i_{j,2}(s-l[i-1]) 
	+ \theta^i_{j,1},
\end{align*}
with \( \Delta l = l[i]-l[i-1] \) and $j\in\left\lbrace x, y \right\rbrace$.

This model has two structural advantages over parameterizing $r$ by polynomials directly. First, the reconstructed curve is smooth by construction, as it is the solution of the differential equation~\eqref{eq:moving_frame} driven by the curvatures. Second, the arc length is controlled exactly: integrating \eqref{eq:moving_frame} only up to a chosen upper bound fixes the total length of the reconstruction, so the known backbone length $L$ is enforced as a hard constraint rather than being approximated through the optimization.

\paragraph*{Remark 1} The moving-frame model is a generalization of the commonly used piecewise constant curvature model, which is obtained by using constants instead of polynomials. We emphasize that this does not impose a physics-based model: the curvatures are free decision variables fitted purely by optimization, and the kinematic constraints of \eqref{eq:moving_frame} are satisfied by construction through the parametrization, so the problem remains unconstrained. Constant curvature is thus a special case of our parametrization not pursued further.\
\paragraph*{Remark 2} The chosen $r$ is a 1D curve in 3D, a valid approximation only for slender robots such as CTCRs, where length greatly exceeds width.
For robots with a higher aspect ratio, adding offsets along the perpendicular directions $R(s)\bolde_1$, $R(s)\bolde_2$ to the tangent $R(s)\bolde_3$ can encode robot-specific features such as width and may improve convergence.

\subsection{Linear projection and distortion}\label{sec:cam}
\noindent
With the reconstruction curve specified, what remains is to make the projection function $T_\gamma$ from Section~\ref{sec:Probform} explicit.
It maps 3D points to pixel coordinates of the $\gamma$-th image in three steps: a pinhole projection from homogeneous world coordinates to normalized camera coordinates, lens distortion (radial and tangential), and finally a scaling and shift to pixel units.

The homogeneous 3D world coordinates \( \mathbf{X_\mathrm{hom}^\mathrm{W}} = [X, Y, Z, 1]^T \) computed from the 3D point \( \mathbf{X}^\mathrm{W} = [X, Y, Z]^T \) give, via linear transformation, the 3D coordinates \( \mathbf{X}^\mathrm{C} = [X^\mathrm{C}, Y^\mathrm{C}, Z^\mathrm{C}]^T \) in the $\gamma$-th camera coordinates with  
\begin{align*}
	\mathbf{X}^\mathrm{C}
	= f^\mathrm{cam}_\gamma(\mathbf{X^\mathrm{W}}) = \begin{bmatrix}
		\mathcal{R}^\gamma& \boldsymbol{t}^\gamma
	\end{bmatrix}
	\begin{bmatrix}\mathbf{X}^\mathrm{W}\\1\end{bmatrix}.
\end{align*}
The projection to normalized camera coordinates is calculated by dividing the vector by its $z$-entry and cutting off the resulting 1 in the third entry with
\begin{align*}
	\mathbf{x}^\mathrm{n} = f^\mathrm{norm}(\mathbf{X}^\mathrm{C}) = \frac{1}{Z^\mathrm{C}}\begin{bmatrix}
		1 & 0 & 0\\
		0 & 1 & 0 
	\end{bmatrix}\mathbf{X}^\mathrm{C}.
\end{align*}
Imperfections in camera lenses cause incorrect projection through, for example, radial and tangential distortion.
Due to radial distortion, straight lines appear curved in images, with increasing effect towards the edges.
It is corrected by identifying the radial distortion coefficients \( k^\gamma_1 \), \( k^\gamma_2 \) and \( k^\gamma_3 \), which are part of the lens distortion model.
Tangential distortion, on the other hand, arises when the camera lens and image plane are not parallel, leading to skewed images.
It is accounted for via the coefficients \( p^\gamma_1 \) and \( p^\gamma_2 \).
These distortion coefficients are determined through the camera calibration process.
With \(
	\mathbf{x}^\mathrm{n} = \begin{bmatrix}
		x^\mathrm{n} & y^\mathrm{n}
	\end{bmatrix}^T\),
the function to distort the image -- like the $\gamma$-th camera's lens does -- is given by \cite{bradski2008learning}
\begin{align*}
	\mathbf{x}^\mathrm{dist} &= f_\gamma^\mathrm{dist}(\mathbf{x}^\mathrm{n})\\ &= \begin{bmatrix}
		x^\mathrm{n} \Phi + p^\gamma_1 \left(2 x^\mathrm{n} y^\mathrm{n}\right)+2 p^\gamma_2 \left(\normexp{\mathbf{x}} + 2 {x^\mathrm{n}}^2\right) \\[5mm]
		y^\mathrm{n} \Phi + 2 p^\gamma_1 \left(\normexp{\mathbf{x}} + 2 {y^\mathrm{n}}^2\right) + p^\gamma_2 \left(2 x^\mathrm{n} y^\mathrm{n}\right)
	\end{bmatrix}\\
	\text{with} \;\Phi &=  \left(1+k^\gamma_1 \normexp{\mathbf{x}^\mathrm{n}} + k^\gamma_2 \norm{\mathbf{x}^\mathrm{n}}^4 + k^\gamma_3 \norm{\mathbf{x}^\mathrm{n}}^6\right).
\end{align*}
Finally, the units are converted to pixels with the camera's intrinsic matrix $\mathcal{K}^\gamma$, containing the focal length and optical center, with
\(
	\mathbf{x} = f_\gamma^\mathrm{pix}(\mathrm{\mathbf{x}^\mathrm{dist}}) =  \mathcal{K}^\gamma \mathbf{x}^\mathrm{dist}.
\)
Composing these yields the continuously differentiable transformation
\begin{align*}
	T_\gamma(\mathbf{X}^\mathrm{W}) = \left(f_\gamma^\mathrm{pix} \circ f_\gamma^\mathrm{dist} \circ f^\mathrm{norm} \circ f_\gamma^\mathrm{cam}\right)(\mathbf{X}^\mathrm{W}).
\end{align*}

\subsection{Initialization strategies}
The gradient-based iterations require an initial $\tilde\theta_0$. We use a straight line, $\theta^i_j = 0$ in \eqref{eq:params}; despite being a poor geometric guess, the next section shows that it leads to satisfactory convergence in most cases.
Better initializations exist: methods such as Camarillo et al. shape-from-silhouette \cite{camarillo2008vision} operate directly on the images without solving an optimization problem, and we expect fitting $\theta$ to the resulting voxel grid provides a warm start that reduces the number of closest-point evaluations and therefore the overall computational cost.
Alternatively, having estimates from a physics-based model can also provide initial guesses, potentially improving convergence and robustness to local minima.


\section{Numerical results}
\label{sec:results_optimal}
\noindent
Building on the algorithms developed in Section~\ref{sec:optimal}, this section assesses their numerical behavior. We first specify the implementation details of Algorithms \ref{met:ms} and \ref{met:os}, ellaborate on the data generation process, and finally compare both variants of our approach with each other and the \os variant with the differentiable-rendering method of Lu et al.\ \cite{lu2023diffrendering}. We single out this method because, among the approaches surveyed in Table~\ref{tab:reconstr_methods}, it is the only one that shares the full scope of ours: stereo (two-camera) input, a continuous marker-free 3D backbone, and no requirement for prior knowledge.
Validation on real images follows in Section~\ref{sec:validation}. For all experiments the moving-frame ODE \eqref{eq:moving_frame} is discretized at 40 equidistant arc lengths, and -- in order to compare runtimes fairly with the PyTorch3D-based differentiable-rendering implementation suggested in \cite{lu2023diffrendering} -- both algorithms are implemented in Python.

The \os algorithm is implemented in PyTorch \cite{paszke2017pytorch}, chosen for its optimizer library and the ease with which the cost function can be updated between iterations; among the available solvers, Adam \cite{kingma2014adam} converged fastest and most reliably. We use stochastic mini-batch gradient descent with batches of 4000 pixels sampled from both images, and integrate the moving frame with the RK4 routine from \mcode{torchdiffeq}~\cite{torchdiffeq}.

The \ms algorithm, in contrast, is formulated in CasADi \cite{andersson2019casadi} and solved with IPOPT \cite{wachter2006ipopt}, and the ODE is integrated by a RK4 routine. 
After each outer iteration a fresh random batch of one fourth of all robot pixels is sampled, so that \ms is stochastic as well.
After convergence, the ODE is re-integrated at 1000 equidistant arc lengths with the adaptive Dormand-Prince \mcode{dopri5} solver for evaluation. All runtimes were measured on the same notebook with a modern Intel processor.

\subsection{Data generation and experimental setting}
\label{sec:data_generation}
\noindent
Ground-truth backbones for comparison are generated with our implementation of the CTCR model of Rucker et al.\,\cite{rucker2010basemodel}, which describes concentrically arranged pre-curved tubes by minimizing the total bending and torsion energy. The resulting boundary-value problem yields the torsion $u_z$ for each tube; $u_x$ and $u_y$ follow from an algebraic relation, and 3D positions are obtained by integrating the moving frame \eqref{eq:moving_frame}.


From these ground-truth backbones we generate artificial images by two complementary pipelines, used in different experiments below.

\subsubsection{Dilation-based image synthesis}
\label{sec:img_dilation}
\begin{figure*}
	\centering
	\begin{subfigure}{.45\textwidth}
		\centering
		\includegraphics[width=0.7\linewidth, trim={10cm 10cm 0cm 4cm}, clip]{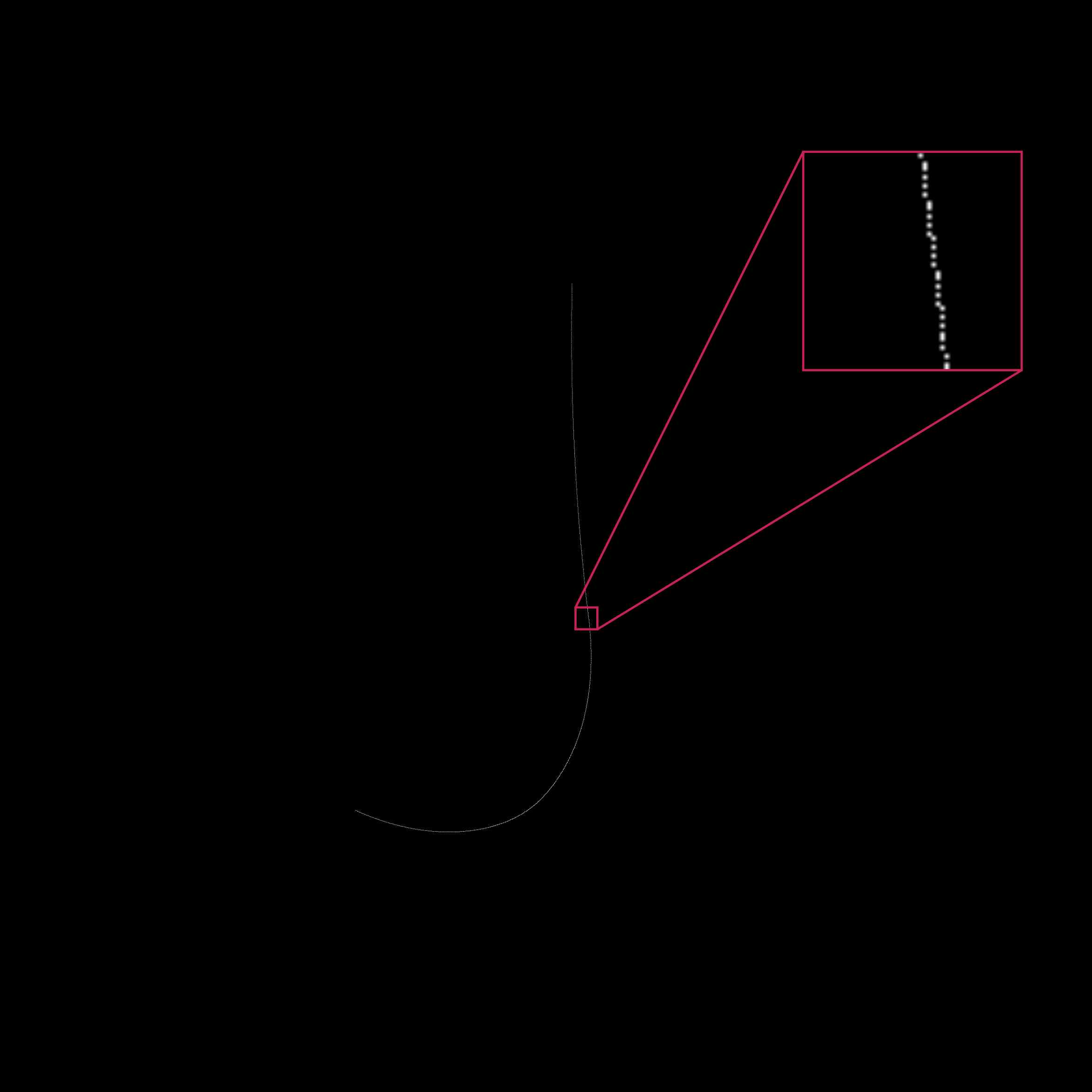}
		\caption{The model by Rucker et al.\,\cite{rucker2010basemodel} is evaluated for 3D positional data. These are projected to the image. After rounding to integers the resulting pixel values, a curve with a thickness of one pixel can be seen in the image.}
		\label{fig:img_base}
	\end{subfigure}%
	\hfill
	\begin{subfigure}{.45\textwidth}
		\centering
		\includegraphics[width=0.7\linewidth, trim={10cm 11cm 1cm 4cm}, clip]{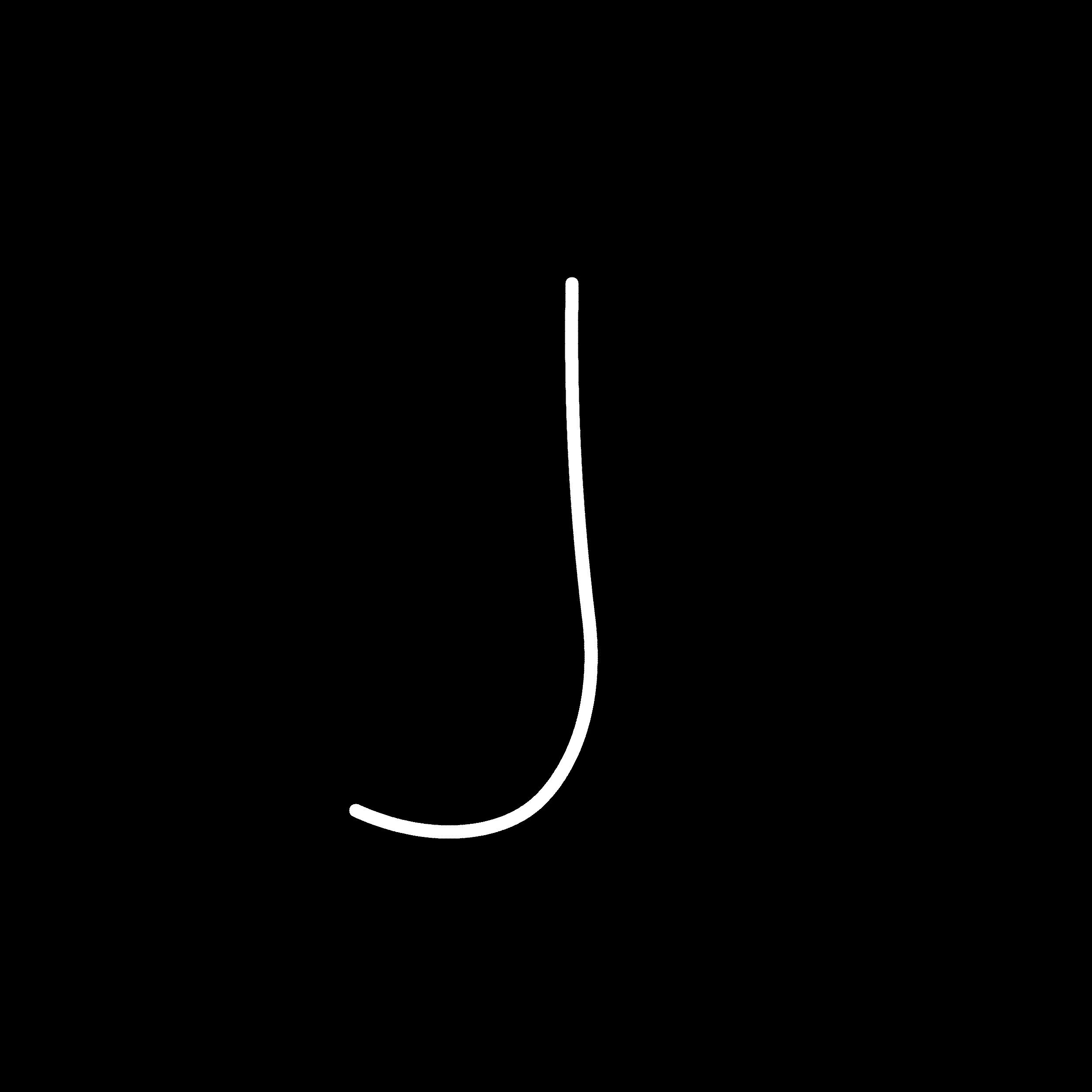}
		\caption{After dilation, the tube has a visible thickness to emulate the volume of the CTCR. \MKHcorrection{A comparison with \figref{fig:binarization of images} shows the similarity with the edited real images.}\\}
		\label{fig:img_morph}
	\end{subfigure}
	\caption{Generation of artificial images by the dilation-based pipeline of Section~\ref{sec:img_dilation}.}
	\label{fig:artif_img}
\end{figure*}

\noindent
The simpler pipeline projects the ground-truth backbone with $T_\gamma$ onto a black canvas, rounds the floating-point image coordinates to pixels, and dilates the resulting one-pixel curve with a disk of radius 15 to give the cannula a realistic thickness \cite{forsyth2003cv}; see Figure~\ref{fig:artif_img}. The procedure approximates the tube silhouette without explicitly modelling its 3D body and produces binary images directly comparable to those obtained from the real test stand after binarisation. This pipeline is used in the reproducibility study of Section~\ref{sec:reproducibility} and in the $p$-value sweep of Section~\ref{sec:choice_of_p}.

\subsubsection{Mesh-based image synthesis with PyTorch3D}
\label{sec:img_mesh}
\noindent
For the head-to-head comparison with differentiable rendering (Section~\ref{sec:comp_diffrendering}) we use a second pipeline: from the ground-truth backbone and the tube radii we construct a triangular mesh of the CTCR and render it in PyTorch3D \cite{ravi2020pytorch3d} using the calibrated camera intrinsics and extrinsics of the scene. Because the differentiable-rendering reconstruction itself operates on this kind of rendered image, comparing both methods on mesh renders eliminates pipeline mismatch between data and reconstruction as a confounding factor in the runtime and error comparison. Example renders are shown alongside the reconstructed Bézier-curve scene and a reconstruction from our approach in Appendix~\ref{app:comparison}, Figure~\ref{fig:comparison_images}.

Error measures are given in absolute terms (mm) and relative to the robot length (\%).

\medskip
\noindent
For both pipelines, all algorithms use the same straight-line initial guess: zero curvatures for the \os and \ms algorithms, and equidistantly spaced collinear Bézier control points for differentiable rendering. Errors are measured as 3D Euclidean distances along the backbone -- evaluated at matching arc lengths in Sections~\ref{sec:reproducibility} and \ref{sec:choice_of_p} (where simulation and reconstruction share an arc-length parametrization) and as nearest-neighbour distances otherwise.

\hyphenation{PyTorch}
\subsection{Comparison with differentiable rendering -- \os algorithm}
\label{sec:comp_diffrendering}
We begin the experiments with a head-to-head comparison against the differentiable rendering pipeline of Lu et al.\ \cite{lu2023diffrendering}, which we re-implemented in PyTorch3D \cite{ravi2020pytorch3d}. The robot's tube radii are assumed to be known and each tube segment is encoded with a three-control-point Bézier curve. Unlike reported in \cite{lu2023diffrendering}, single-image reconstruction did not succeed for our CTCRs; we therefore use two images and sum the per-image cost functions. The total cost combines the mask, keypoint, distance, and appearance terms of \cite[equations (10)--(14)]{lu2023diffrendering}. Although this adaptation of the original single-image pipeline is modest, it is essential: without the second view, differentiable rendering did not converge to a comparable reconstruction for any of our CTCR configurations.

The artificial images for this experiment are produced by the mesh-based pipeline of Section~\ref{sec:img_mesh}: a 3-tube CTCR extending in the $z$-direction (parameters in Appendix \ref{app:robot_params} Table~\ref{tab:tube_params_2}) is meshed and rendered through PyTorch3D from two cameras placed on the $x$- and $y$-axes at \SI{90}{\degree}. Rotating the CTCR around its $z$-axis in \SI{10}{\degree} steps yields 36 poses in total.

\figref{fig:diffrender_icp_comparison} shows the mean and maximum deviation over the full backbone from the ground truth of both reconstruction methods, either via differentiable rendering (DiffRender) or ours (ICP).
We find that our method is able to generate estimations with errors on average below \SI{1}{\milli\meter} / \SI{0.5}{\percent} and maximum deviations of around \SI{1}{\milli\meter} in between 14 and \SI{18}{\second}. 
Exceptionally, for angles of \SI{60}{\degree} and \SI{330}{\degree}, the reconstruction enters the tube in the image from the tip, similar to the problem reported in the next subsection in \figref{fig:match_in_images} for the \ms algorithm.
The exact images and the faulty reconstruction are placed in the Appendix in \figref{fig:bad_convergence}.
Self-occlusion as it appears for angles \SI{70}{\degree}, \SI{160}{\degree}, \SI{250}{\degree} and \SI{340}{\degree} does not pose a problem, since the self-occlusion only appears in one of the two images.

Differentiable rendering, in contrast, produces average errors of at least \SI{4}{\milli\meter} / \SI{2}{\percent} and converges in 108--\SI{113}{\second}. We attribute the gap to two factors: chains of Bézier curves lack the smoothness and length hard-constraints of the moving frame, increasing the degrees of freedom. Moreover, the distance map underlying their distance loss corresponds to the \(R\)-to-\(B\) correspondence with its continuum of local minima.
\def\fileone{data/error_over_angle_icp.csv}
\def\filetwo{data/error_over_angle_spline.csv}

\begin{figure}
\centering
\begin{tikzpicture}
	\begin{axis}[
		axis x line=bottom,
		xlabel=Angle,
		ylabel=Error (mm),
		ymode=log,
		xmin=0, xmax=360,
		ymin=0,
		legend style={at={(0.5,-0.22)},anchor=north, /tikz/every even column/.append style={column sep=1cm}},
		legend columns=2,
		ymajorgrids,
		xmajorgrids,
		scaled x ticks = false,
		axis y line=left,
		enlargelimits=0.05,
		xtick align=inside,
		ytick align=inside,
		axis on top,
		axis line style={-},
		width=0.6*\textwidth,
		]
		
		\addplot[
		color=udsblue,
		solid,
		mark options={solid},
		line width=1.5pt,
		] table[x=Angle, y expr=\thisrow{Mean GT to Model Error}*1000, col sep=comma] {\fileone};
		\addlegendentry{Mean Error (ICP)};
		
		\addplot[
		color=udsblue,
		dashed,
		mark options={solid},
		line width=1.5pt,
		] table[x=Angle, y expr=\thisrow{Mean GT to Model Error}*1000, col sep=comma] {\filetwo};
		\addlegendentry{Mean Error (DiffRender)};
		
		\addplot[
		color=udsred,
		solid,
		mark options={solid},
		line width=1.5pt,
		] table[x=Angle, y expr=\thisrow{Max GT to Model Error}*1000, col sep=comma] {\fileone};
		\addlegendentry{Max Error (ICP)};
		
		\addplot[
		color=udsred,
		dashed,
		mark options={solid},
		line width=1.5pt,
		] table[x=Angle, y expr=\thisrow{Max GT to Model Error}*1000, col sep=comma] {\filetwo};
		\addlegendentry{Max Error (DiffRender)};
		
	\end{axis}
	
%
%
%
\end{tikzpicture}
\caption{Comparison of our approach (ICP) with differential rendering-based reconstruction (DiffRender) \cite{lu2023diffrendering} on the same dataset of 36 rendered images of a single CTCR configuration, rotated in \SI{10}{\degree} steps.
We find that our ICP-based approach consistently outperforms the differential rendering-based approach in both mean and maximum error for CTCRs.}
\label{fig:diffrender_icp_comparison}
\end{figure}

A side-by-side comparison of the rendered Bézier-curve scene with the target image after convergence of the differentiable rendering algorithm at an angle of \SI{40}{\degree} is given in Appendix \ref{app:comparison}, Figure~\ref{fig:comparison_images}.

While the Python implementation already reduces the convergence time by an order of magnitude, \SI{17}{\second} per pose is still too slow for real-time control. A comparable implementation of the \os algorithm in Rust reduces this gap, completing the same task in \SI{0.534}{\second}.

\subsection{Reproducibility -- \os and \ms algorithm}
\label{sec:reproducibility}
\noindent
Having established that the \os algorithm outperforms differentiable rendering for CTCRs, we next turn to the two variants of our own method and assess their reproducibility. Since both \os and \ms are stochastic, we repeat 50 reconstructions of the same configuration (model as described in Appendix \ref{app:robot_model} and Table~\ref{tab:tube_params}) and record the maximum Euclidean distance between ground truth and reconstruction at matching arc lengths; the distributions are shown as boxplots in Figure~\ref{fig:boxplots_base}. The \os algorithm with step length $\alpha = 0.2$ achieves an average maximum error of \SI{0.665}{\milli\meter} / \SI{0.3325}{\percent} (worst case \SI{1.368}{\milli\meter} / \SI{0.684}{\percent}, best case \SI{0.225}{\milli\meter} / \SI{0.1125}{\percent}) in about \SI{17}{\second}, while \ms converges in \SI{18}{\second} but with low cost variance to an incorrect reconstruction, yielding an average maximum distance of \SI{30.02}{\milli\meter} / \SI{15.01}{\percent}.

\begin{figure*}
	\centering
	\begin{tikzpicture}
	\begin{groupplot}[
		group style={
			group size=2 by 1, 
			horizontal sep=2cm, 
		},
		boxplot/draw direction=y,
		height=6cm, 
		]
		
		\nextgroupplot[
		ylabel={Maximum distance in \SI{}{\milli\meter}},
		xtick={1, 2, 3},
		xticklabels={\os},
		xticklabel style={rotate=45, anchor=east},
		width=4cm,
		]
		\addplot [
		boxplot,
		color=udsblue,
		line width=1.pt,
		] table [y index=0, y expr=\thisrowno{0}*0.001] {data/stochastic_dist_torch.txt};
		
		
		
		\nextgroupplot[
		ylabel={},
		xtick={1},
		xticklabels={\ms},
		xticklabel style={rotate=45, anchor=east},
		width=4cm,
		]
		
		\addplot[
		boxplot,
		color=udsred,
		line width=1.pt,
		] table [y index=0, y expr=\thisrowno{0}*0.001] {data/stochastic_dist_casadi.txt};
		
	\end{groupplot}
\end{tikzpicture}
	\caption{The \os algorithm and \ms algorithm with \(p=8\) fit the reconstruction to the pixel with maximum distance to ground truth of around \SI{1}{\milli\meter}, whereas the \ms version with \(p=2\) converges to a local minimum.}
	\label{fig:boxplots_base}
\end{figure*}
 
Figure~\ref{fig:match_in_images} shows representative reconstructions and the initial guess: the \os algorithm with $p=2$ captures the shape accurately, while \ms is trapped in a physically meaningless local minimum where the reconstruction leaves the tube and re-enters at the tip.
\begin{figure*}
	\centering
	\input{figs/match_in_image}
	\caption{The \os algorithm fits the reconstruction to the pixel, whereas the \ms algorithm with $p=2$ converges to a local minimum.}
	\label{fig:match_in_images}
\end{figure*}

We attribute the gap to the different update behavior of the closest-point correspondences, in particular the late switching of \ms discussed in Section~\ref{sec:Probform}.

\subsection{Choice of $p$ value -- \os and \ms algorithm}
\label{sec:choice_of_p}
The previous experiment used a single fixed configuration. To check that it was representative of the larger configuration space and to study how the norm exponent $p$ in the distance computation affects accuracy, we sweep both algorithms across 10,645 configurations of a 2-tube CTCR (Unit 1 and 2), varying curvatures, tube extension and tube rotation. \figref{fig:histogram_configs} shows the maximum-error histograms on a logarithmic count axis, and Table~\ref{tab:reconstruction_errors} summarizes the fraction of runs below \SI{1}{\milli\meter} and \SI{2}{\milli\meter}. The \os algorithm dominates \ms in all cases, and its accuracy improves monotonically with $p$, whereas \ms shows no such trend. For $p=8$, \os keeps the maximum error below \SI{1}{\milli\meter} in \SI{96.1}{\percent} of cases and below \SI{2}{\milli\meter} (smaller than the outer tube diameter) in \SI{99.3}{\percent}. The worst-case maximum error for \os decreases from \SI{41}{\milli\meter} ($p=2$) to \SI{15}{\milli\meter} ($p=8$), while for \ms it ranges between \SI{77}{\milli\meter} and \SI{130}{\milli\meter} with no systematic trend. Remaining failures are the same tip-re-entry local minimum seen for \ms in \figref{fig:match_in_images}; this case is easy to detect automatically by checking the maximum deviation between reconstruction and robot pixels.
\begin{figure}
	\centering

\begin{tikzpicture}

\definecolor{darkgray176}{RGB}{176,176,176}
\definecolor{steelblue31119180}{RGB}{31,119,180}

\pgfmathsetmacro{\shift}{0.48}

\begin{groupplot}[
	group style={
		group size=2 by 2,
		vertical sep=1.5cm,
		horizontal sep=2cm,
	},
	width=0.48\textwidth,
	ymode=log,
	xlabel={Maximum error in \SI{}{\milli\meter}},
	ylabel={Count},
	xticklabel style={/pgf/number format/fixed},
	enlarge x limits=3e-2,
	ymajorgrids=false,
	xmajorgrids=false,
	axis on top,
	log origin y=infty,
	xmax = 140
]

\nextgroupplot[
	title={$p=2$},
	legend style={at={(0.5,-0.15)}, anchor=north, legend columns=-1, column sep=1.5mm},
	legend cell align={left},
	legend image code/.code={\draw [#1] (0.1cm,-0.15cm) rectangle (0.2cm,0.15cm);},
	bar width=0.48,
	xlabel={},
]
\addplot+[
	ybar,
	fill=udsblue!50,
	draw=udsblue,
	x filter/.code={\pgfmathparse{\pgfmathresult-\shift}},
] table [
	x=bincenter,
	y=count,
	col sep=comma,
	mark=none,
] {data/p_norm_parameter_study/MaxErrorHistogramData_os2.csv};

\addplot+[
	ybar,
	fill=udsred!50,
	draw=udsred,
	x filter/.code={\pgfmathparse{\pgfmathresult+\shift}},
] table [
	x=bincenter,
	y=count,
	col sep=comma,
	mark=none,
] {data/p_norm_parameter_study/MaxErrorHistogramData_ms2.csv};

\nextgroupplot[
	title={$p=4$},
	legend style={at={(0.5,-0.15)}, anchor=north, legend columns=-1, column sep=1.5mm},
	legend cell align={left},
	legend image code/.code={\draw [#1] (0.1cm,-0.15cm) rectangle (0.2cm,0.15cm);},
	bar width=0.48,
	xlabel={},
	ylabel={}
]
\addplot+[
	ybar,
	fill=udsblue!50,
	draw=udsblue,
	x filter/.code={\pgfmathparse{\pgfmathresult-\shift}},
] table [
	x=bincenter,
	y=count,
	col sep=comma,
	mark=none,
] {data/p_norm_parameter_study/MaxErrorHistogramData_os4.csv};

\addplot+[
	ybar,
	fill=udsred!50,
	draw=udsred,
	x filter/.code={\pgfmathparse{\pgfmathresult+\shift}},
] table [
	x=bincenter,
	y=count,
	col sep=comma,
	mark=none,
] {data/p_norm_parameter_study/MaxErrorHistogramData_ms4.csv};

\nextgroupplot[
	title={$p=6$},
	legend style={at={(1.2,-0.3)}, anchor=north, legend columns=-1, column sep=1.5mm},
	legend cell align={left},
	legend image code/.code={\draw [#1] (0.1cm,-0.15cm) rectangle (0.2cm,0.15cm);},
	bar width=0.48,
]
\addplot+[
	ybar,
	fill=udsblue!50,
	draw=udsblue,
	x filter/.code={\pgfmathparse{\pgfmathresult-\shift}},
] table [
	x=bincenter,
	y=count,
	col sep=comma,
	mark=none,
] {data/p_norm_parameter_study/MaxErrorHistogramData_os6.csv};
\addlegendentry{\os~~~~~}

\addplot+[
	ybar,
	fill=udsred!50,
	draw=udsred,
	x filter/.code={\pgfmathparse{\pgfmathresult+\shift}},
] table [
	x=bincenter,
	y=count,
	col sep=comma,
	mark=none,
] {data/p_norm_parameter_study/MaxErrorHistogramData_ms6.csv};
\addlegendentry{\ms}

\nextgroupplot[
	title={$p=8$},
	legend style={at={(0.5,-0.15)}, anchor=north, legend columns=-1, column sep=1.5mm},
	legend cell align={left},
	legend image code/.code={\draw [#1] (0.1cm,-0.15cm) rectangle (0.2cm,0.15cm);},
	bar width=0.48,,
	ylabel={},
]
\addplot+[
	ybar,
	fill=udsblue!50,
	draw=udsblue,
	x filter/.code={\pgfmathparse{\pgfmathresult-\shift}},
] table [
	x=bincenter,
	y=count,
	col sep=comma,
	mark=none,
] {data/p_norm_parameter_study/MaxErrorHistogramData_os8.csv};

\addplot+[
	ybar,
	fill=udsred!50,
	draw=udsred,
	x filter/.code={\pgfmathparse{\pgfmathresult+\shift}},
] table [
	x=bincenter,
	y=count,
	col sep=comma,
	mark=none,
] {data/p_norm_parameter_study/MaxErrorHistogramData_ms8.csv};

\end{groupplot}
\end{tikzpicture}
	\caption{Histograms with logarithmic count axis of the maximum error along the reconstructed backbones of 10,645 simulated CTCRs.
	The \os algorithm outperforms \ms in all cases, and its worst-case error decreases monotonically with $p$:
	\SI{41}{\milli\meter}, \SI{31}{\milli\meter}, \SI{28}{\milli\meter}, \SI{15}{\milli\meter} for $p=2,4,6,8$.
	For \ms the worst-case error shows no systematic trend:
	\SI{77}{\milli\meter}, \SI{117}{\milli\meter}, \SI{130}{\milli\meter}, \SI{94}{\milli\meter} for $p=2,4,6,8$.}
	\label{fig:histogram_configs}
\end{figure}

To make the errors comparable across robots of different size, \figref{fig:histogram_configs_rel} repeats the analysis with the maximum error expressed relative to the robot length. The relative errors confirm the absolute-error picture: the \os algorithm keeps the worst-case relative error below \SI{20}{\percent} for every $p$, whereas for \ms it grows monotonically from \SI{307}{\percent} at $p=2$ to \SI{458}{\percent} at $p=8$. Normalizing by robot length therefore shows that the accuracy gap between the two algorithms is not an artefact of a particular robot scale but persists across the whole configuration space. The median relative error for \os is \SI{0.27}{\percent} ($p=2$), \SI{0.26}{\percent} ($p=4$), \SI{0.24}{\percent} ($p=6$), and \SI{0.23}{\percent} ($p=8$), confirming that the typical reconstruction lies well below \SI{1}{\percent} across the full configuration space. For \ms the median is \SI{0.59}{\percent}, \SI{0.76}{\percent}, \SI{1.06}{\percent}, and \SI{1.24}{\percent} for $p=2,4,6,8$.
\begin{figure}
	\centering

\begin{tikzpicture}

\definecolor{darkgray176}{RGB}{176,176,176}
\definecolor{steelblue31119180}{RGB}{31,119,180}

\pgfmathsetmacro{\shift}{2}

\begin{groupplot}[
	group style={
		group size=2 by 2,
		vertical sep=1.5cm,
		horizontal sep=2cm,
	},
	width=0.48\textwidth,
	ymode=log,
	xlabel={Maximum relative error in \SI{}{\percent}},
	ylabel={Count},
	xticklabel style={/pgf/number format/fixed},
	enlarge x limits=3e-2,
	ymajorgrids=false,
	xmajorgrids=false,
	axis on top,
	log origin y=infty,
	xmax = 500
]

\nextgroupplot[
	title={$p=2$},
	legend style={at={(0.5,-0.15)}, anchor=north, legend columns=-1, column sep=1.5mm},
	legend cell align={left},
	legend image code/.code={\draw [#1] (0.1cm,-0.15cm) rectangle (0.2cm,0.15cm);},
	bar width=2,
	xlabel={},
]
\addplot+[
	ybar,
	fill=udsblue!50,
	draw=udsblue,
	x filter/.code={\pgfmathparse{\pgfmathresult-\shift}},
] table [
	x=bincenter,
	y=count,
	col sep=comma,
	mark=none,
] {data/p_norm_parameter_study/MaxRelativeErrorHistogramData_os2.csv};

\addplot+[
	ybar,
	fill=udsred!50,
	draw=udsred,
	x filter/.code={\pgfmathparse{\pgfmathresult+\shift}},
] table [
	x=bincenter,
	y=count,
	col sep=comma,
	mark=none,
] {data/p_norm_parameter_study/MaxRelativeErrorHistogramData_ms2.csv};

\nextgroupplot[
	title={$p=4$},
	legend style={at={(0.5,-0.15)}, anchor=north, legend columns=-1, column sep=1.5mm},
	legend cell align={left},
	legend image code/.code={\draw [#1] (0.1cm,-0.15cm) rectangle (0.2cm,0.15cm);},
	bar width=2,
	xlabel={},
	ylabel={}
]
\addplot+[
	ybar,
	fill=udsblue!50,
	draw=udsblue,
	x filter/.code={\pgfmathparse{\pgfmathresult-\shift}},
] table [
	x=bincenter,
	y=count,
	col sep=comma,
	mark=none,
] {data/p_norm_parameter_study/MaxRelativeErrorHistogramData_os4.csv};

\addplot+[
	ybar,
	fill=udsred!50,
	draw=udsred,
	x filter/.code={\pgfmathparse{\pgfmathresult+\shift}},
] table [
	x=bincenter,
	y=count,
	col sep=comma,
	mark=none,
] {data/p_norm_parameter_study/MaxRelativeErrorHistogramData_ms4.csv};

\nextgroupplot[
	title={$p=6$},
	legend style={at={(1.2,-0.3)}, anchor=north, legend columns=-1, column sep=1.5mm},
	legend cell align={left},
	legend image code/.code={\draw [#1] (0.1cm,-0.15cm) rectangle (0.2cm,0.15cm);},
	bar width=2,
]
\addplot+[
	ybar,
	fill=udsblue!50,
	draw=udsblue,
	x filter/.code={\pgfmathparse{\pgfmathresult-\shift}},
] table [
	x=bincenter,
	y=count,
	col sep=comma,
	mark=none,
] {data/p_norm_parameter_study/MaxRelativeErrorHistogramData_os6.csv};
\addlegendentry{\os~~~~~}

\addplot+[
	ybar,
	fill=udsred!50,
	draw=udsred,
	x filter/.code={\pgfmathparse{\pgfmathresult+\shift}},
] table [
	x=bincenter,
	y=count,
	col sep=comma,
	mark=none,
] {data/p_norm_parameter_study/MaxRelativeErrorHistogramData_ms6.csv};
\addlegendentry{\ms}

\nextgroupplot[
	title={$p=8$},
	legend style={at={(0.5,-0.15)}, anchor=north, legend columns=-1, column sep=1.5mm},
	legend cell align={left},
	legend image code/.code={\draw [#1] (0.1cm,-0.15cm) rectangle (0.2cm,0.15cm);},
	bar width=2,,
	ylabel={},
]
\addplot+[
	ybar,
	fill=udsblue!50,
	draw=udsblue,
	x filter/.code={\pgfmathparse{\pgfmathresult-\shift}},
] table [
	x=bincenter,
	y=count,
	col sep=comma,
	mark=none,
] {data/p_norm_parameter_study/MaxRelativeErrorHistogramData_os8.csv};

\addplot+[
	ybar,
	fill=udsred!50,
	draw=udsred,
	x filter/.code={\pgfmathparse{\pgfmathresult+\shift}},
] table [
	x=bincenter,
	y=count,
	col sep=comma,
	mark=none,
] {data/p_norm_parameter_study/MaxRelativeErrorHistogramData_ms8.csv};

\end{groupplot}
\end{tikzpicture}
	\caption{Histograms with logarithmic count axis of the maximum relative error along the reconstructed backbones of 10,645 simulated CTCRs.
	The \os algorithm outperforms \ms in all cases, and its worst-case relative error remains below \SI{20}{\percent}:
	\SI{19}{\percent}, \SI{15}{\percent}, \SI{17}{\percent}, \SI{17}{\percent} for $p=2,4,6,8$.
	For \ms the worst-case relative error shows a continuously increasing trend:
	\SI{307}{\percent}, \SI{407}{\percent}, \SI{440}{\percent}, \SI{458}{\percent} for $p=2,4,6,8$.}
	\label{fig:histogram_configs_rel}
\end{figure}

\begin{table}[htb]
	\centering
	\caption{Percent of algorithm runs resulting in reconstruction errors (in mm) less than 1 and \SI{2}{\milli\meter} for different values of $p$ using the \os and \ms algorithms.}
	\begin{tabular}{|c|c|cccc|}\hline
		& less than & $p=2$ & $p=4$ & $p=6$ & $p=8$ \\
		\hline
		\os& \SI{1}{\milli\meter} & \SI{85.7}{\percent} & \SI{86.1}{\percent} & \SI{91.9}{\percent} & \SI{96.1}{\percent} \\
		\ms& \SI{1}{\milli\meter} & \SI{54.8}{\percent} & \SI{54.1}{\percent} & \SI{52.1}{\percent} & \SI{51.1}{\percent} \\
		\os& \SI{2}{\milli\meter} & \SI{92.8}{\percent} & \SI{92.7}{\percent} & \SI{96.3}{\percent} & \SI{99.3}{\percent} \\
		\ms& \SI{2}{\milli\meter} & \SI{58.7}{\percent} & \SI{57.1}{\percent} & \SI{54.7}{\percent} & \SI{54.0}{\percent} \\\hline
	\end{tabular}
	\label{tab:reconstruction_errors}
\end{table}

All numerical experiments so far assume perfectly modelled image projections; camera imperfections (cf.\ Section~\ref{sec:cam}) in real-world settings will degrade any image-based reconstruction, including ours. The next section therefore moves from artificial images to a real CTCR test stand.

\section{Validation on real images}
\label{sec:validation}

This section first describes the hardware and toolchain of a test stand required to acquire high-detail binary images from CRs as an input to the described algorithms. After that, the \os~algorithm is validated on different real poses of a CTCR against manual photogrammetric measurements.
This validation is intended as a proof of concept and for simulating data generation for model validation, rather than a comprehensive accuracy benchmark.

\subsection{Test stand}

Images are acquired on a CTCR test stand with an automated three-tube actuation system and two industrial cameras \cite{Muehlenhoff2022} mounted at \SI{90}{\degree} to each other around the exit-direction axis for an accuracy-optimal layout \cite{Muehlenhoff2024} (see \figref{fig_testbench}). Camera sensor resolution is chosen against tube thickness and accuracy demands; monochrome rolling-shutter $5\,\mathrm{MP}$ sensors with a telephoto lens of matched optical resolution proved sufficient. Apertures of about $f/8$ balance depth-of-field and diffraction blur, and low sensor gain yields exposure times of $1$--$5\,\text{ms}$ depending on lighting. Backlight illumination from a monochromatic LED panel maximizes contrast and avoids chromatic aberration.

\begin{figure}
	\centering
	\input{figs/testbench.tex}
	\caption{Layout of the test stand with the CTCR and its actuation system and the arrangement of the two industry cameras and two LED panels.}
	\label{fig_testbench}
\end{figure}

\subsection{Image processing}

\begin{figure}
	\centering
	\begin{subfigure}[t]{0.32\textwidth}
		\centering
		\includegraphics[width=0.9\textwidth]{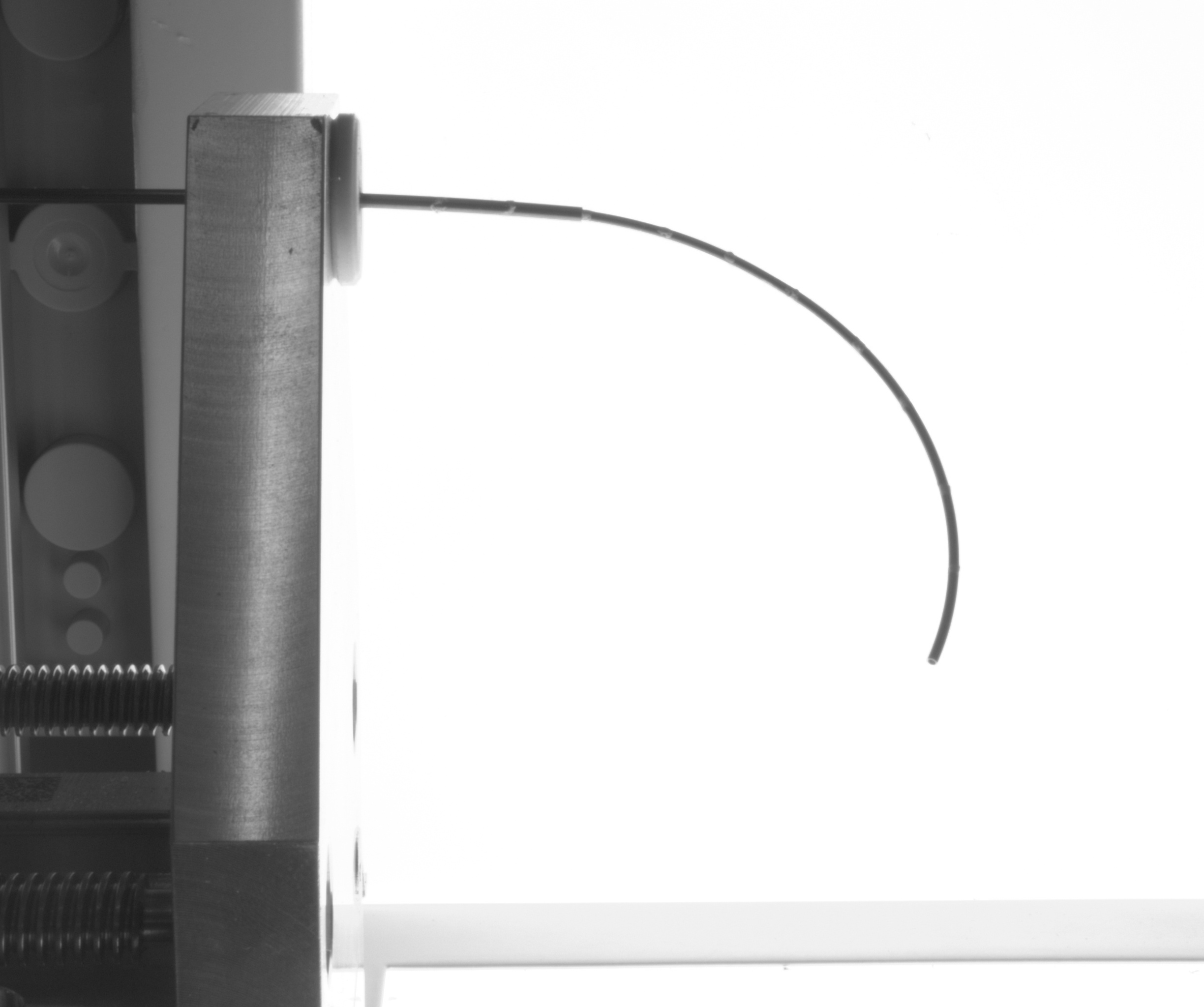}\\
		\vspace{0.2cm}
		\includegraphics[width=0.9\textwidth]{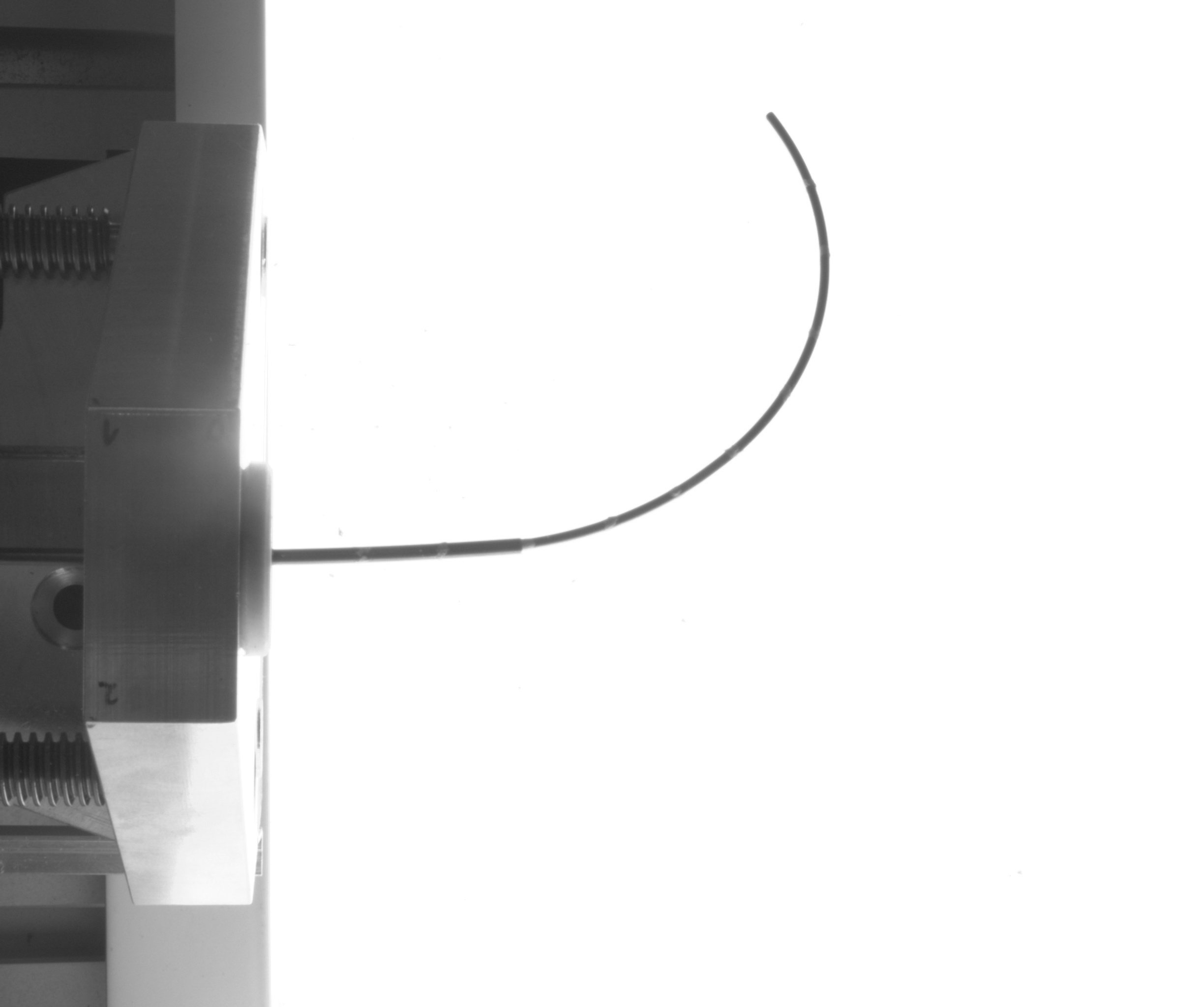}
		\caption{Raw images captured from cameras with parts of the test stand still visible.\\}
		\label{fig:real images captured from cameras}
	\end{subfigure}
	\hfill
	\begin{subfigure}[t]{0.32\textwidth}
		\centering
		\includegraphics[width=0.9\textwidth]{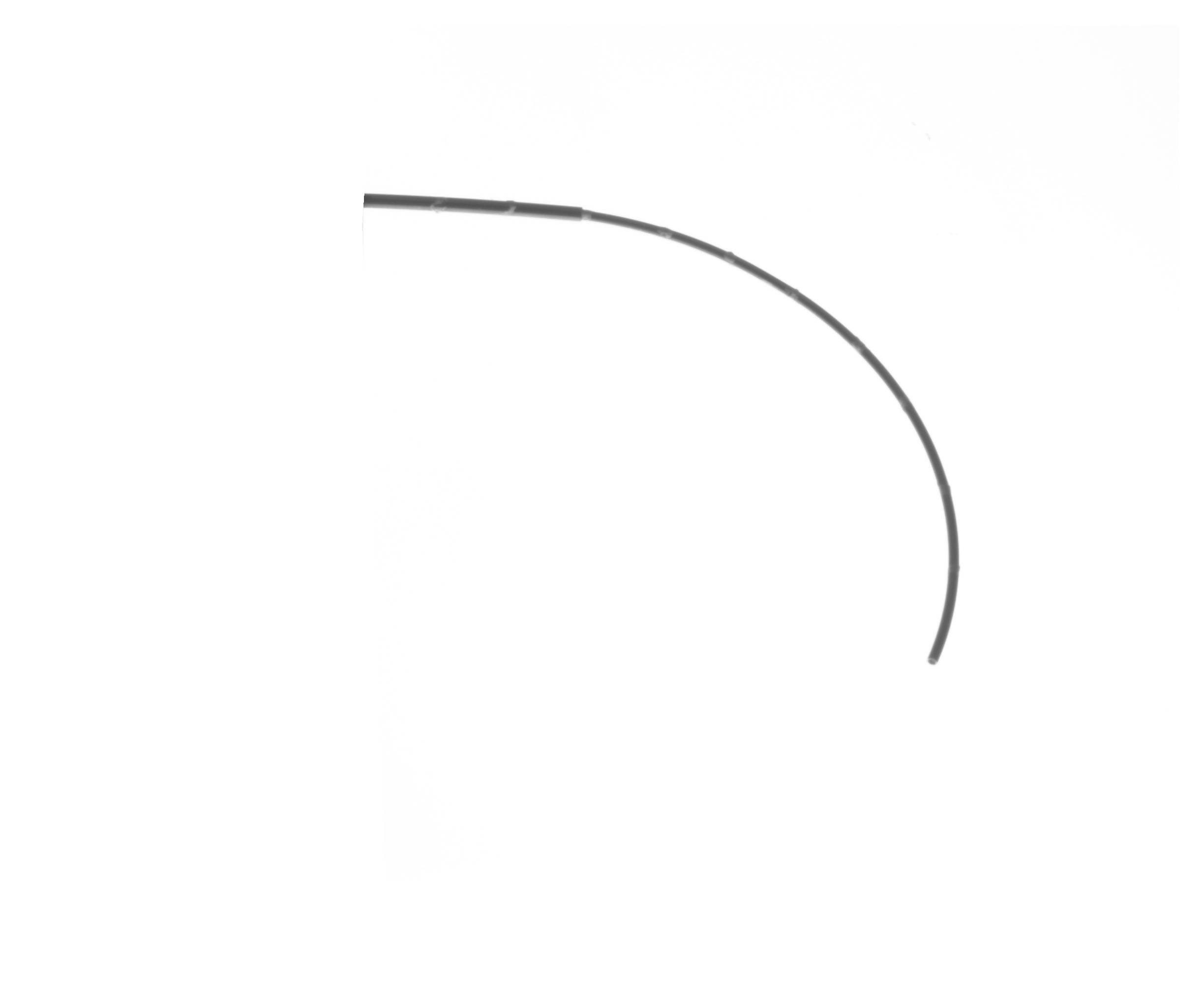}\\
		\vspace{0.2cm}
		\includegraphics[width=0.9\textwidth]{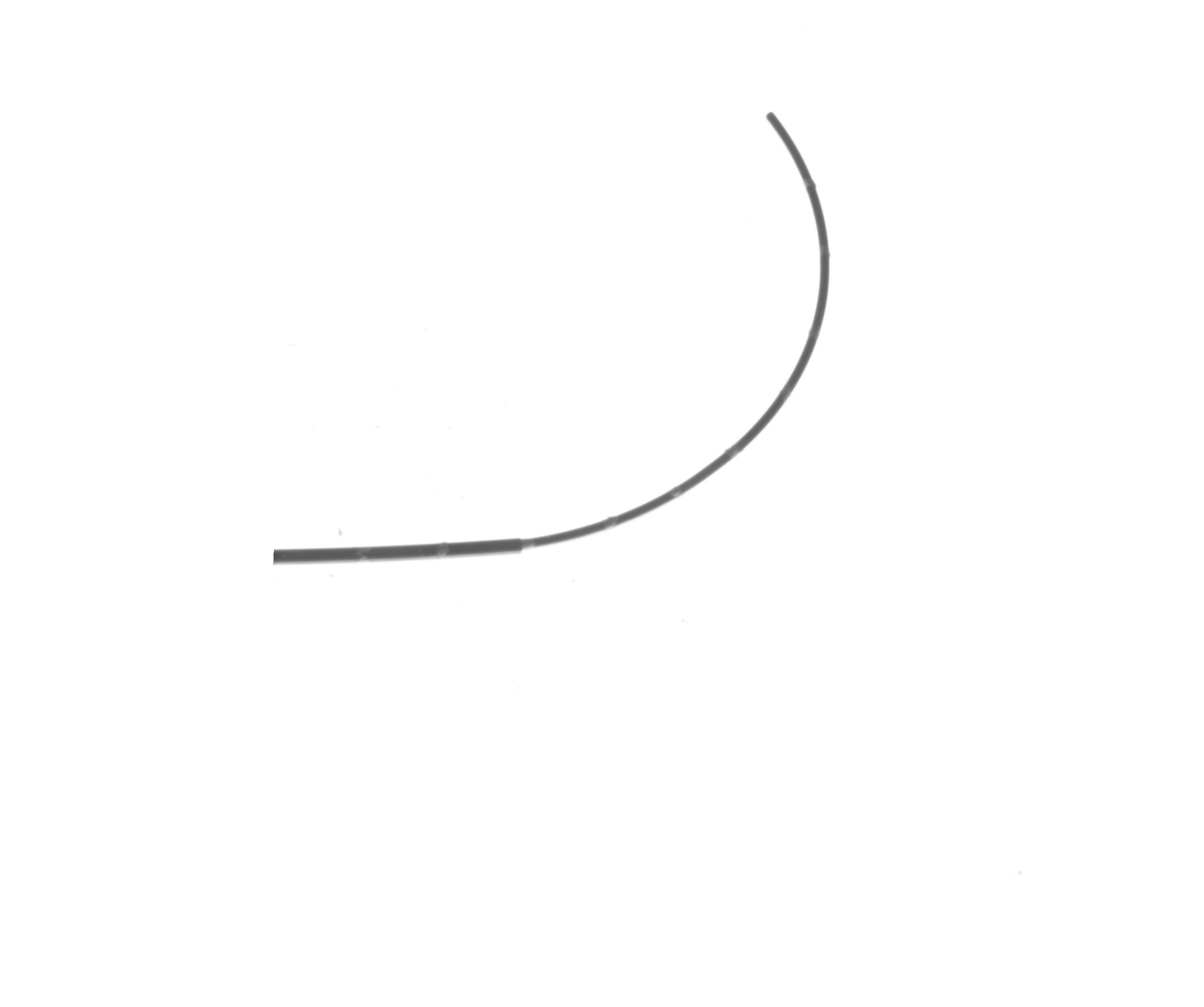}
		\caption{Image cropping via static cutoff regions.\\}
		\label{fig:image cropping and noise filtering}
	\end{subfigure}
	\hfill
	\begin{subfigure}[t]{0.32\textwidth}
		\centering
		\includegraphics[width=0.9\textwidth]{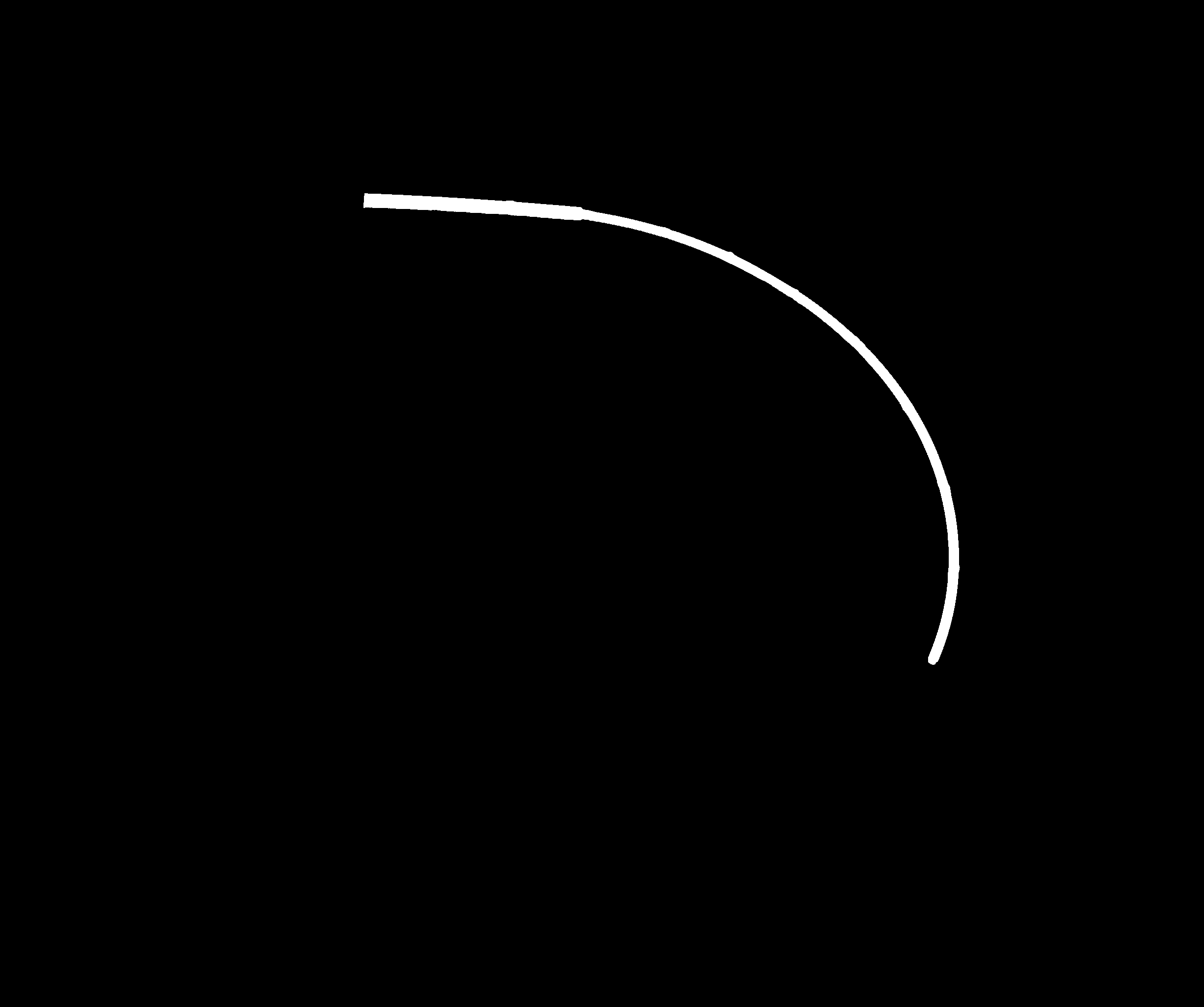}\\
		\vspace{0.2cm}
		\includegraphics[width=0.9\textwidth]{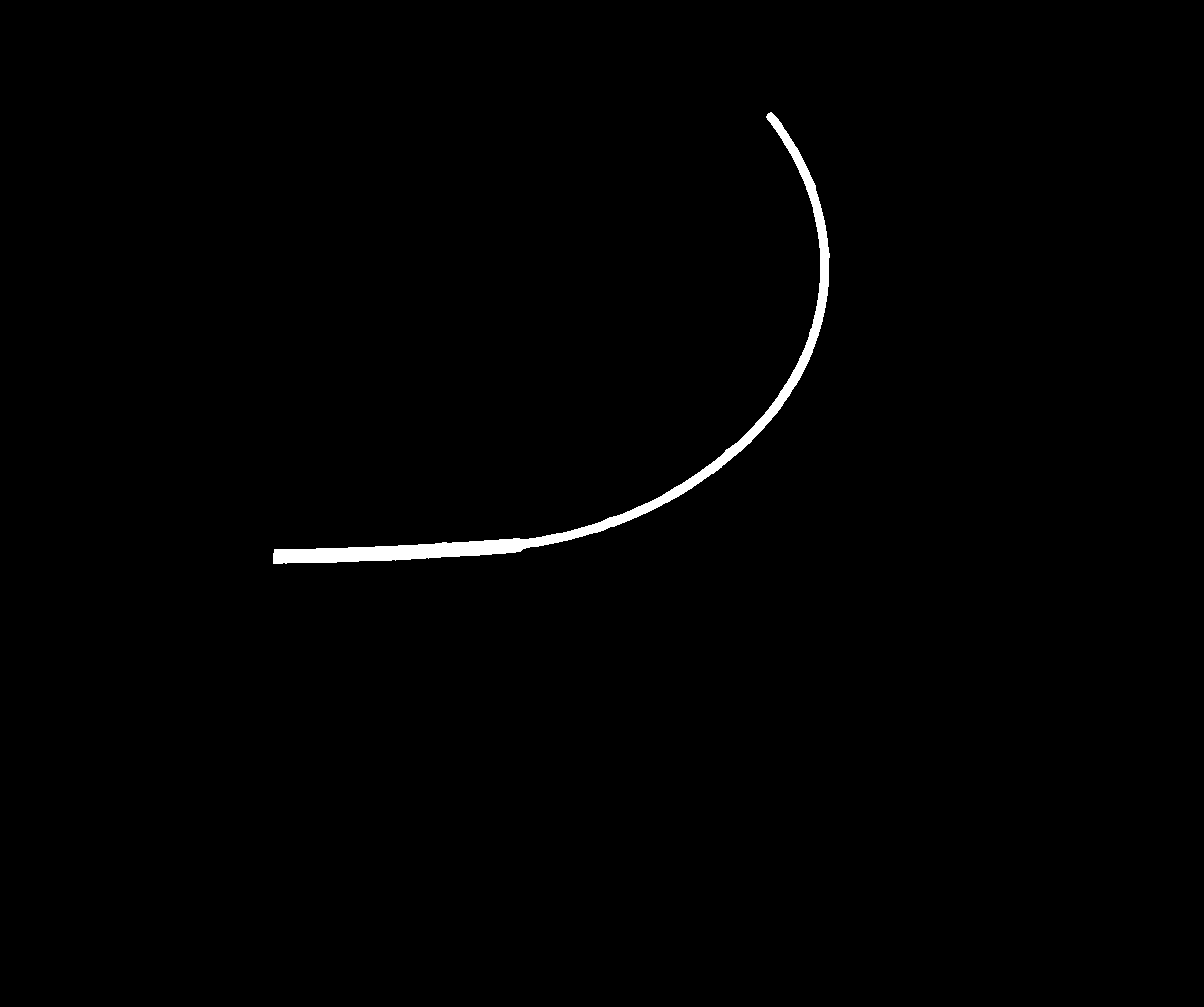}
		\caption{Binarization of images via thresholding and noise filtering by applying morphological opening.}
		\label{fig:binarization of images}
	\end{subfigure}
	\caption{Image processing workflow from the camera's raw image to the final binary image used for reconstruction.}
	\label{fig_image_processing_examples}
\end{figure}

Image processing aims at a noise-free binary image of the entire backbone from each camera. Static foreground elements (e.g.\ the actuation system's exit point) that remain visible despite backlighting are masked out using a one-time manual region selection valid for the whole test-stand setup. The masked image is then thresholded to a binary image; residual noise is removed by morphological opening, and the image is inverted for convenience. \figref{fig_image_processing_examples} shows the full workflow on one exemplary pose.

\subsection{Validation}

\begin{figure*}
	\centering	
	\begin{subfigure}[t]{.3\textwidth}
		\centering
		\includegraphics[width=1.0\linewidth]{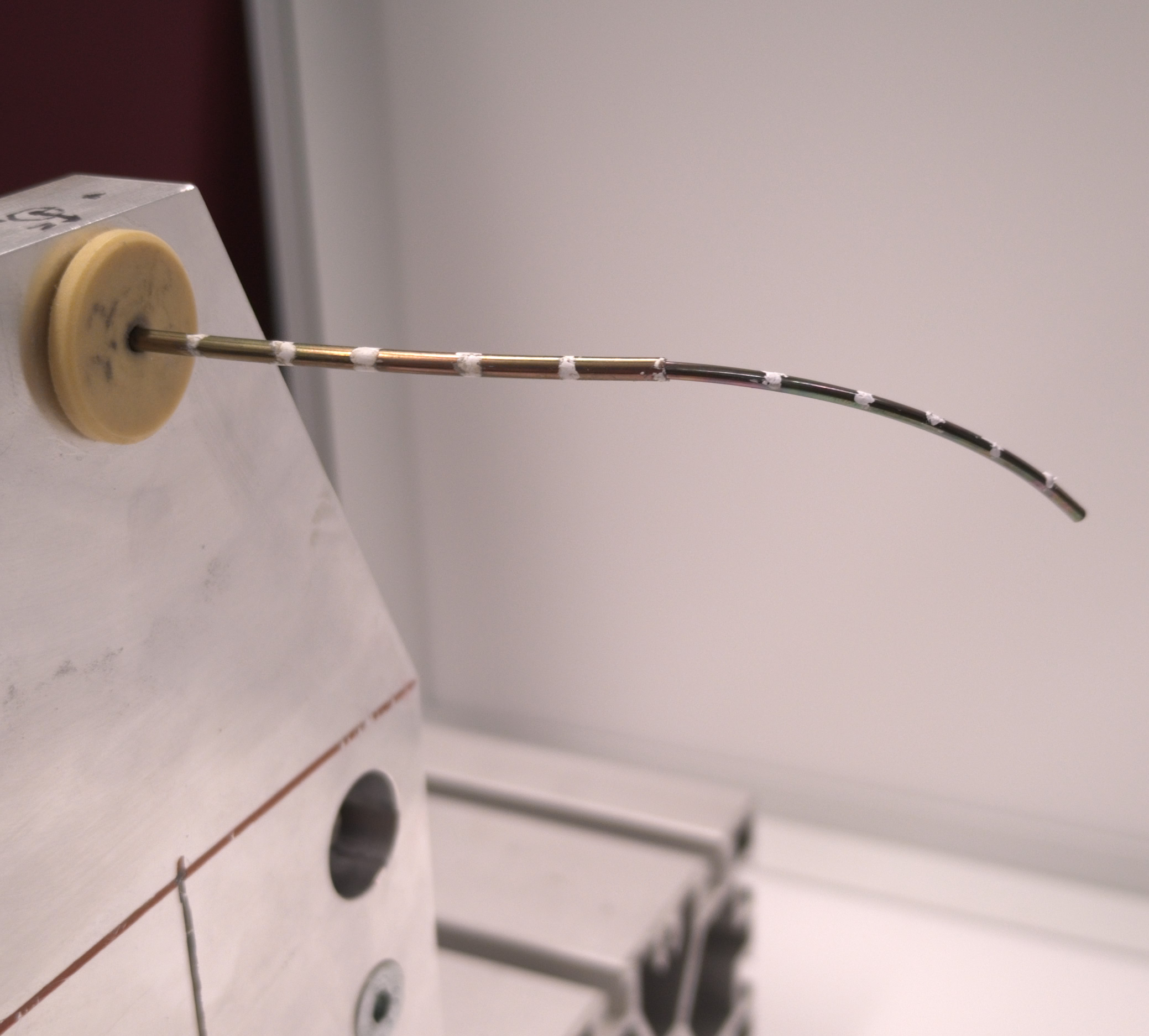}
		\caption{Two-tube-CTCR with its annotation markers for manual search of corresponding image points, showing only one exemplary robot pose.}
		\label{fig:real image of CTCR}
	\end{subfigure}%
	\hfill
	\begin{subfigure}[t]{.33\textwidth}
		\centering
		\vbox{\vfill \begin{tikzpicture}
	\small
	
	\definecolor{darkorange25512714}{RGB}{255,127,14}
	
	\begin{axis}[
		axis equal image,
		axis lines=none,
		width=7.3cm,
		view={25}{18},
		unit vector ratio*={1 1 1},
		xmin= -0.01, xmax=0.05,
		ymin=-0.04, ymax=0.02,
		zmin=-0, zmax=0.08,
		clip=false,
		ticks=none,
		legend style={legend entries={Proj. onto coord. planes, Manual measurement, \os algorithm},
			legend pos = north west, xshift = -6.5mm, yshift=9.5mm, font = \small,  legend cell align=left},
		rotate around y= 90,
		rotate around z= 90,
		xlabel={$X$}, 
		xlabel style={at={(axis cs:0.025,-0.029,0)}, anchor=north east, xshift = 1mm},
		ylabel={$Y$}, 
		ylabel style={at={(axis cs:0.01,-0.01,-0.01)}, anchor=north east, rotate=-90},
		zlabel={$Z$},
		zlabel style={rotate=-90}
		]		

		\addplot3 [color=gray, thin] table[col sep=space, x index=0, y index=1, z index=2] {data/Data_3d_op_new_z.txt};
		
		\addplot3 [color=gray, thin, forget plot] table[col sep=space, x index=0, y index=1, z index=2] {data/Data_3d_op_new_y.txt};
		
		\addplot3 [color=gray, thin, forget plot] table[col sep=space, x index=0, y index=1, z index=2] {data/Data_3d_op_new_x.txt};
		
		
		\addplot3 [mark=*, only marks, color=udsred, semithick] table[col sep=space, x index=0, y index=1, z index=2] {data/Data_3d_manuell_new.txt};
		
	
		\addplot3 [color=udsblue, thick] table[col sep=space, x index=0, y index=1, z index=2] {data/Data_3d_op_new.txt};
		
		\addplot3 [black, thin, forget plot] coordinates {(-0.01,-0.04,0)    ( 0.05,-0.04,0)};
		\addplot3 [black, thin, forget plot] coordinates {(-0.01, 0.02,0)    ( 0.05, 0.02,0)};
		\addplot3 [black, thin, forget plot] coordinates {(-0.01,-0.04,0.08) ( 0.05,-0.04,0.08)};
		\addplot3 [black, thin, forget plot] coordinates {(-0.01, 0.02,0.08) ( 0.05, 0.02,0.08)};
		\addplot3 [black, thin, forget plot] coordinates {(-0.01,-0.04,0)    (-0.01, 0.02,0)};
		\addplot3 [black, thin, forget plot] coordinates {( 0.05,-0.04,0)    ( 0.05, 0.02,0)};
		\addplot3 [black, thin, forget plot] coordinates {(-0.01,-0.04,0.08) (-0.01, 0.02,0.08)};
		\addplot3 [black, thin, forget plot] coordinates {( 0.05,-0.04,0.08) ( 0.05, 0.02,0.08)};
		\addplot3 [black, thin, forget plot] coordinates {(-0.01,-0.04,0)    (-0.01,-0.04,0.08)};
		\addplot3 [black, thin, forget plot] coordinates {(-0.01, 0.02,0)    (-0.01, 0.02,0.08)};
		\addplot3 [black, thin, forget plot] coordinates {( 0.05,-0.04,0)    ( 0.05,-0.04,0.08)};
		\addplot3 [black, thin, forget plot] coordinates {( 0.05, 0.02,0)    ( 0.05, 0.02,0.08)};
		
		
	\end{axis}
	
\end{tikzpicture}}
		\caption{Validation of the One-Step algorithm against the manually triangulated backbone points, showing the robot pose of Figure~\ref{fig:real image of CTCR}.}
		\label{fig:tube_epipolar_3d}
	\end{subfigure}
	\hfill
	\begin{subfigure}[t]{.3\textwidth}
		\centering
		\vbox{\vfill 

\begin{tikzpicture}
	\small
	\begin{axis}[
		boxplot/draw direction=y,
		height=5.5cm, 
		width=3.5cm,
		ylabel={Max. distance in \SI{}{\milli\meter}},
		xtick=\empty,
		]
		
		\addplot [
		boxplot,
		color=udsblue,
		line width=1.pt,
		] table [y index=0, y expr=\thisrowno{0}] {data/stochastic_readist_2step_new.txt};
		
	\end{axis}
\end{tikzpicture}}
		\caption{Deviation between the 3D shape estimated using the \os{} algorithm and manual triangulation, based on all 18 poses and 50 algorithm runs per pose.}
		\label{fig:boxplots_validation_realimage}
	\end{subfigure}
	\caption{Validation through real images.}
	\label{fig:Validation through real image}
\end{figure*}

To validate the \os algorithm on real images, we annotated 10 points on the backbone of each of two CTCRs with white markers (Figure~\ref{fig:real image of CTCR}); the CTCR parameters are given in Table~\ref{tab:tube_params_3}.

\begin{table}[!h]
	\small
	\renewcommand{\arraystretch}{1.2}
	\centering
	\begin{tabular}{| c | c |c | c | c | c | c | c |}
		\hline
		CTCR & Tube & $d_\mathrm{i}$ & $d_\mathrm{o}$ & $u_x^\star$ & $u_y^\star$ & $u_z^\star$ & $l$\\ \hline
		& & \SI{}{\milli\meter} & \SI{}{\milli\meter} & \SI{}{\per\milli\meter} & \SI{}{\per\milli\meter} & \SI{}{\radian\per\milli\meter} & \SI{}{\milli\meter}\\ \hline
		\multirow[c]{2}{*}{1}& 1 & 0 & 1.6 & 20 & 0 & 0 & 100 \\ \cline{2-8}
		& 2 & 1.8 & 2.1 & 12.5 & 0 & 0 & 50\\ \hline
		\multirow[c]{2}{*}{2}& 1 & 0 & 1.6 & 10 & 0 & 0 & 100 \\ \cline{2-8}
		& 2 & 1.8 & 2.1 & 6.25 & 0 & 0 & 50\\ \hline
	\end{tabular}
	\caption{Tube parameters of two different CTCRs used for validation with real images, see section \ref{sec:validation}.}
	\label{tab:tube_params_3}
\end{table}

For each pose, image positions of corresponding markers were selected manually in both camera views and triangulated to 3D, providing the reference against which the reconstruction is compared. The procedure was repeated for 9 poses per CTCR (18 in total), evenly sampled across the rotatory actuation space, and therefore including out-of-plane curvature for non-trivial configurations, where the bending-axes are not aligned (c.f.~projections in Figure~\ref{fig:tube_epipolar_3d}). Because the tubes are actuated by a combination of rotation and translation, the resulting configurations are genuinely spatial: they combine bending and twisting rather than remaining planar. We stress that the aim of this validation is not to demonstrate generalizability across robot types -- CTCRs are inherently restricted in the shapes they can assume, and our actuation system further constrains the reachable workspace -- but to verify the algorithm on real, spatially deformed configurations.

Figure~\ref{fig:tube_epipolar_3d} illustrates, for one example pose, that the reconstruction obtained by the \os algorithm stays close to the manually triangulated points. The segment lengths $l[\cdot]$ are read directly from the electrical actuation system; the starting position $\boldsymbol{p}(0)$ is selected manually in each camera image, and the initial orientation $\boldsymbol{R}(0)$ is obtained from reference points on the actuation unit.

To turn this qualitative agreement into a quantitative measure, we report errors on all 18 poses. Because only ten manually marked points are available per pose, the error measure of Section~\ref{sec:results_optimal} -- the largest distance from the reconstruction to the ground-truth backbone -- is inverted here: we instead report the maximum distance from each manual measurement to the closest point on the reconstruction. The \os formulation was run with $p=2$ and step length $\alpha = 0.2$ for 20 epochs, 50 times per pose. Across all 18 poses the algorithm achieved an average maximum distance of \SI{0.903}{\milli\meter} / \SI{0.903}{\percent}, with a worst case of \SI{2.0}{\milli\meter} / \SI{2}{\percent} and a best case of \SI{0.395}{\milli\meter} / \SI{0.395}{\percent}; the full distribution is shown in Figure~\ref{fig:boxplots_validation_realimage}.

\begin{figure*}[htb]
	\centering
\begin{tikzpicture}[scale=0.8]
	
	\definecolor{steelblue31119180}{RGB}{31,119,180}
	
	\begin{groupplot}[group style={group size=2 by 2},
		xticklabels={},
		yticklabels={},]
		
		\nextgroupplot[
		tick pos=left,
		title={Raw Image 0},
		xmin=-0.5, xmax=2448.5,
		y dir=reverse,
		x dir = reverse,
		ymin=-0.5, ymax=2048.5
		]
		\addplot graphics [xmin=-0.5, xmax=2448, ymin=2048, ymax=-0.5] {C0_1.jpg};
		\addplot [line width=1.5, color=udsgreen, semithick] table[col sep=space, x index=0, y index=1] {data/backbone_op_1.txt};
		
		\nextgroupplot[
		tick pos=left,
		title={Raw Image 1},
		xmin=-0.5, xmax=2448.5,
		y dir=reverse,
		ymin=-0.5, ymax=2048.5
		]
		\addplot graphics [xmin=-0.5, xmax=2448, ymin=2048, ymax=-0.5] {C1_1.jpg};
		\addplot [line width=1.5, color=udsgreen, semithick] table[col sep=space, x index=0, y index=1] {data/backbone_op_2.txt};
		
		\nextgroupplot[
		tick pos=left,
		title={Binary Image 0},
		xmin=-0.5, xmax=2448.5,
		y dir=reverse,
		x dir = reverse,
		ymin=-0.5, ymax=2048.5
		]
		\addplot graphics [xmin=-0.5, xmax=2448, ymin=2048, ymax=-0.5] {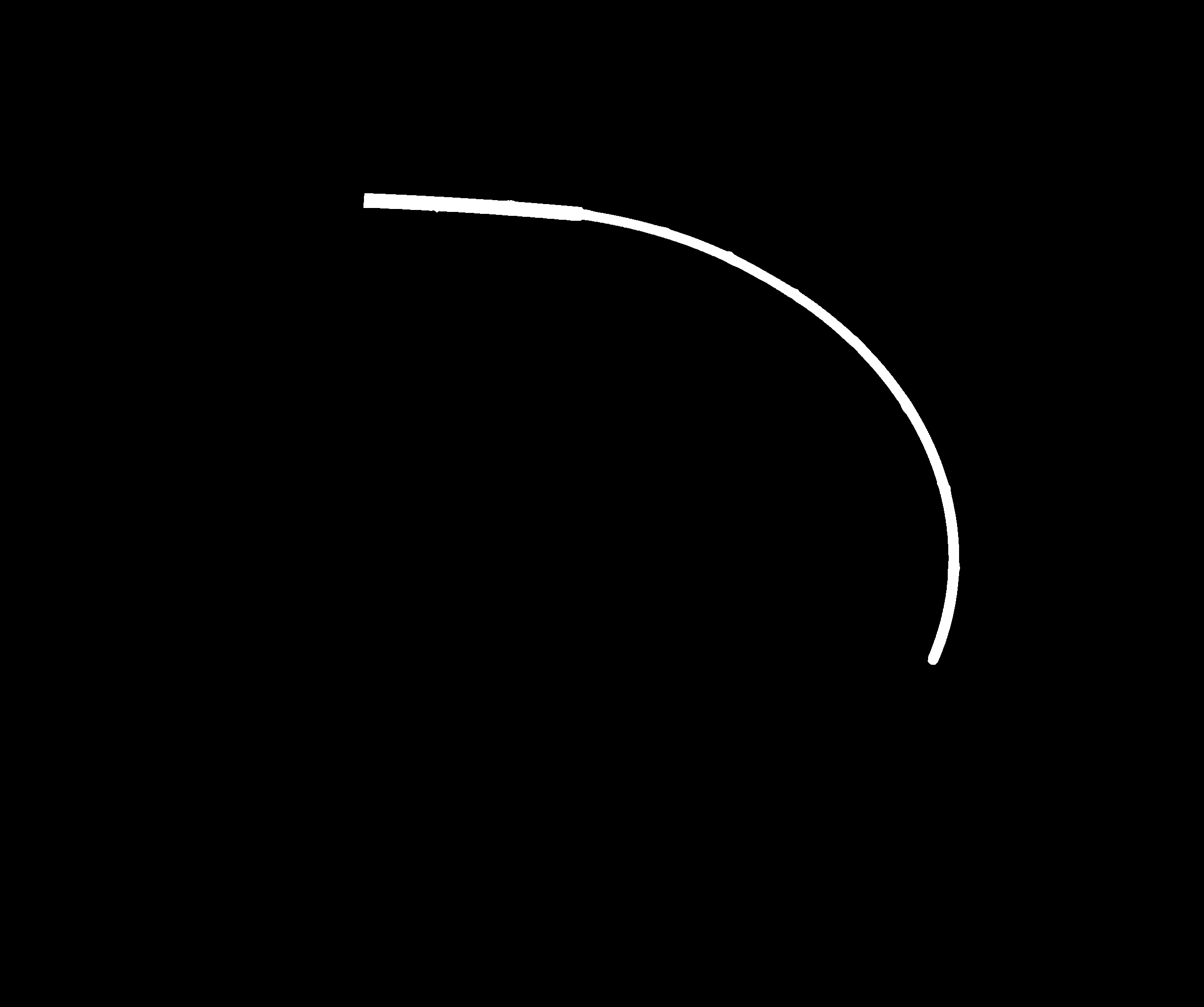};
		\addplot [line width=1.5, color=udsgreen, semithick] table[col sep=space, x index=0, y index=1] {data/backbone_op_1.txt};
		\legend{,Projected reconstruction };
		
		\nextgroupplot[
		tick pos=left,
		title={Binary Image 1},
		xmin=-0.5, xmax=2448,
		y dir=reverse,
		ymin=-0.5, ymax=2048,
		]
		\addplot graphics [xmin=-0.5, xmax=2448, ymin=2048, ymax=-0.5] {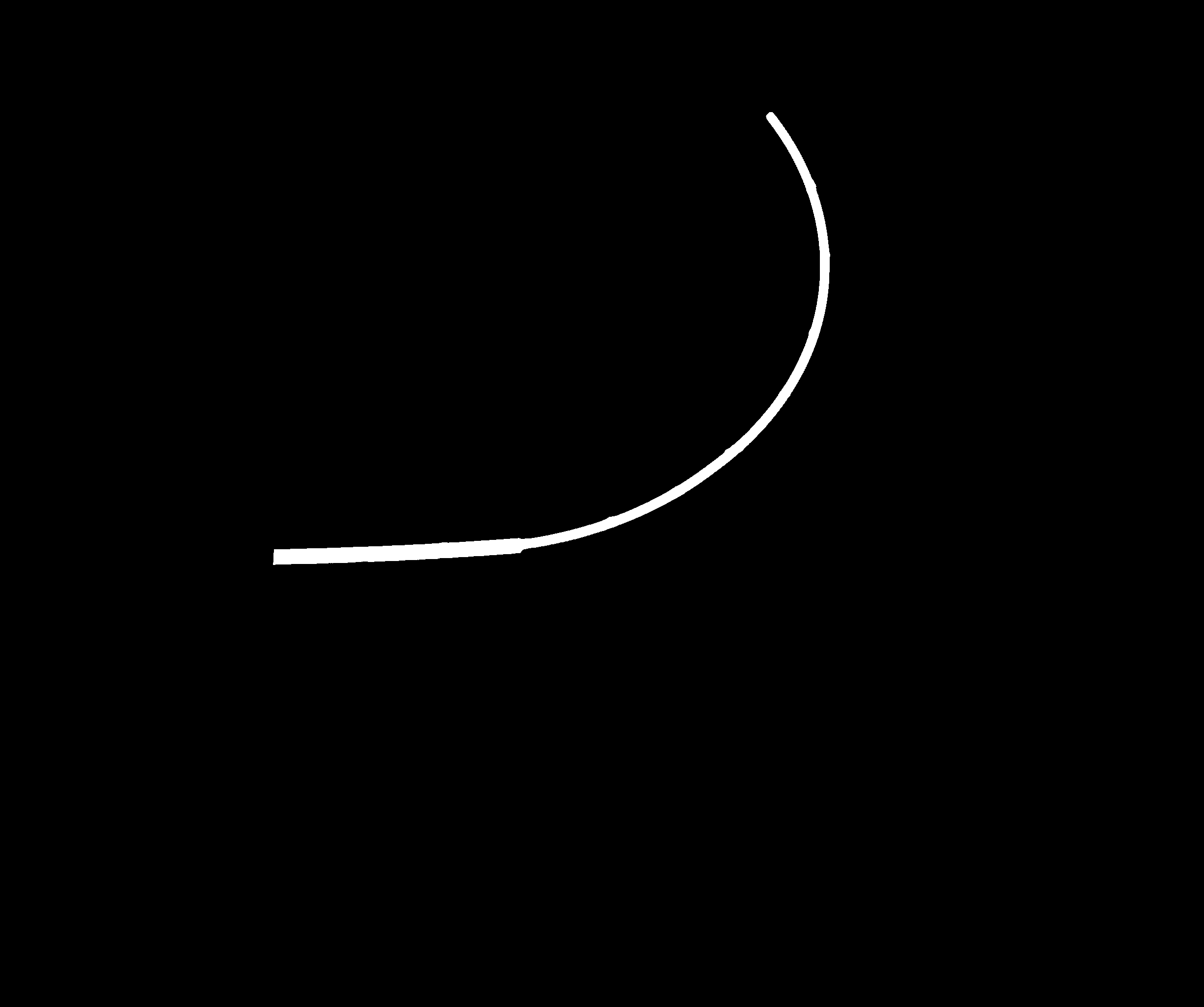};
		\addplot [line width=1.5, color=udsgreen, semithick] table[col sep=space, x index=0, y index=1] {data/backbone_op_2.txt};
		
	\end{groupplot}
	
\end{tikzpicture}
	\caption{Results of the validation, exemplary for one measurement. The projected reconstruction lies inside the binarised concentric tube continuum robot. Compare to \figref{fig_image_processing_examples} for the raw images and binarization steps.}
	\label{fig:backbone_realimage}
\end{figure*}

These errors are larger than for the artificial images, which is expected: the markers and the manual point correspondence definition introduce additional human error sources not present in the simulated setting. Despite this, Figure~\ref{fig:backbone_realimage} shows that the projected reconstruction follows the visible robot shape and stays inside the area occupied by the tubes. The experiment is intended as a proof of concept for the algorithm's basic functionality on real images rather than a comprehensive accuracy benchmark; a broader characterization across different CR designs and operating conditions is left for future work. In line with our scope, the method targets accurate, marker-free shape measurement for model validation and the generation of reference data; transferring it to less controlled scenes would mainly require a more advanced image-processing front-end, while the reconstruction core remains unchanged. A direct quantitative comparison with other shape-estimation techniques is not possible here, as -- to our knowledge -- no published method operates in the same setting (cf.\ Table~\ref{tab:reconstr_methods}): two cameras, marker-free, a full continuous 3D backbone, and no prior knowledge.

\newpage

\section{Summary and Outlook}
\label{sec:summary_outlook}
\noindent
We presented a marker-free, image-based reconstruction method for continuum robots that fits a parametric 3D backbone curve directly to binary image observations, using an iterative closest point formulation that can combine an arbitrary number of camera views. Two variants were proposed: the \ms and the \os formulation. By operating directly on the 1D backbone rather than on a rendered 3D volume, the approach reduces the convergence time of the same reconstruction task from about \SI{108}{\second} for differentiable rendering to roughly \SI{17}{\second} in our Python implementation; a comparable implementation in Rust converges in \SI{0.534}{\second}. At the same time, on artificial images of CTCRs the average backbone error drops from at best \SI{4}{\milli\meter} for differentiable rendering to below \SI{1}{\milli\meter} for our method.

Across the further experiments, the \os formulation showed clear advantages in reproducibility, and higher values of the norm exponent $p$ consistently improved accuracy in comparison to the \ms formulation. Finally, on real images of two CTCRs across 18 poses we observed agreement with manual photogrammetric measurements at an average maximum deviation below \SI{1}{\milli\meter}.

Several directions for future work follow naturally. Warm-starting the algorithm -- either from physics-based models or from image-only methods such as those discussed in Section~\ref{sec_shape_estimation} -- could further reduce the number of correspondence updates and runtime. The framework currently builds on the standard ICP formulation; integrating non-standard ICP variants from the broader literature may improve robustness, in particular against partial occlusions and image-processing artefacts. 
While the 1D backbone parametrization fits slender CTCRs well, thicker or otherwise non-slender continuum robots will require a curve parametrization adapted to specific robots that for example encode the width of the robot.
The static parametrization used here matches the quasi-static actuation of CRs. Extending it to time-dependent parametrizations would capture dynamic effects and open the door to identifying dynamic models, which we regard as a promising direction for future work. While the current implementation is not yet fast enough for real-time tracking, recorded videos can already be processed offline as a batch of frames, making dynamic sequences accessible in post-processing. In a related direction, the moving-frame formulas recover the backbone shape only; the problem of reconstructing the true material strain from the projected shape alone is underdetermined, and adapting the parametrization to estimate them is an interesting extension towards true state estimation.
Finally, extending the method to other classes of continuum robots and to more challenging scenes -- with occlusions and cluttered backgrounds -- remains an open question.

\section*{Funding}
This work was supported by the Deutsche Forschungsgemeinschaft (DFG, German Research Foundation) under Grants 501928699; 461953135.

\section*{Data availability}
The data that support the findings of this study are available from the corresponding author, MKH, upon reasonable request.

\section*{Disclosure of interest}
The authors report there are no competing interests to declare.

\printbibliography

\newpage
\appendix

\section{Robot parameters used for experiments}\label{app:robot_params}

\subsection{Comparison with differentiable rendering}
\noindent
The comparison with differentiable rendering in Section~\ref{sec:comp_diffrendering} uses the sample CTCR specified in Table~\ref{tab:tube_params_2}.

\begin{table}[htbp]
	\small
	\renewcommand{\arraystretch}{1.2}
	\centering
	\begin{tabular}{| c |c | c | c | c | c | c | c | c | c | c |}
		\hline
		Tube & $d_\mathrm{i}$ & $d_\mathrm{o}$ & $E$ & $\nu$ & $u_x^\star$ & $u_y^\star$ & $u_z^\star$ & $L$ & $\alpha$ & $\beta$\\ \hline
		Unit & \SI{}{\milli\meter} & \SI{}{\milli\meter} & \SI{}{\giga\pascal} & 1 & \SI{}{\per\milli\meter} & \SI{}{\per\milli\meter} & \SI{}{\radian\per\milli\meter} & \SI{}{\milli\meter} & \SI{}{\degree} & \SI{}{\milli\meter}\\ \hline
		1 & 1.0 & 1.4 & 58 & 0.3488 & 40 & 0 & 0 & 200 & -45 & -50\\ \hline
		2 & 1.5 & 2.4 & 58 & 0.3488 & 5 & 0 & 0 & 130 & 0 & -30\\ \hline
		3 & 2.5 & 4.0 & 58 & 0.3488 & 15 & 0 & 0 & 70 & 45 & -20\\ \hline
	\end{tabular}
	\caption{Tube parameters used for comparison with the differentiable rendering technique, see section \ref{sec:comp_diffrendering}.}
	\label{tab:tube_params_2}
\end{table}

\subsection{\JMcorrection{Ground truth data}}\label{app:robot_model}
\noindent
The artificial images in Section~\ref{sec:results_optimal} are produced from the CTCR model of Rucker et al.\,\cite{rucker2010basemodel}; the parameters are listed in Table~\ref{tab:tube_params}. Geometric parameters ($d_\mathrm{i}$, $d_\mathrm{o}$, the pre-curvatures $u_{x,y,z}^\star$ and the lengths $L$) correspond to manufactured experimental tubes, the material constants $E$ (Young's modulus) and $\nu$ (Poisson's ratio) are taken from \cite{Rucker2011}, and the actuation values $\alpha$ (rotation) and $\beta$ (translation) were chosen arbitrarily.
\begin{table}[htbp]
	\small
	\renewcommand{\arraystretch}{1.2}
	\centering
	\begin{tabular}{| c |c | c | c | c | c | c | c | c | c | c |}
		\hline
		Tube & $d_\mathrm{i}$ & $d_\mathrm{o}$ & $E$ & $\nu$ & $u_x^\star$ & $u_y^\star$ & $u_z^\star$ & $L$ & $\alpha$ & $\beta$\\ \hline
		Unit & \SI{}{\milli\meter} & \SI{}{\milli\meter} & \SI{}{\giga\pascal} & 1 & \SI{}{\per\milli\meter} & \SI{}{\per\milli\meter} & \SI{}{\radian\per\milli\meter} & \SI{}{\milli\meter} & \SI{}{\radian} & \SI{}{\milli\meter}\\ \hline
		1 & 0 & 1.6 & 58 & 0.3488 & 0 & 0.0238 & 0 & 200 & -1.0 & -10\\ \hline
		2 & 2.01 & 2.39 & 58 & 0.3488 & 0 & 0.0099 & 0 & 140 & 1.2 & -10\\ \hline
		3 & 2.5 & 3.5 & 58 & 0.3488 & 0.005 & 0.0 & 0 & 80 & -0.8 & -5\\ \hline
	\end{tabular}
	\caption{Tube parameters used in the reproducibility tests, see section \ref{sec:reproducibility}.}
	\label{tab:tube_params}
\end{table}


\newpage
\section{Images of algorithm comparison}
\label{app:comparison}
\begin{figure}[!htbp]
	\centering
\begin{tikzpicture}[scale=0.8]
	
	\definecolor{steelblue31119180}{RGB}{31,119,180}
	
	\begin{groupplot}[group style={group size=2 by 2},
		xticklabels={},
		yticklabels={},
		xmin=-0.5, xmax=2448,
		x dir=reverse,
		ymin=-0.5, ymax=2048,]
		\nextgroupplot[
		title={Target Image Camera 0},
		]
		
		\addplot graphics [xmin=-0.5, xmax=2448, ymin=2048, ymax=-0.5] {comparison_diff_render/curve_mesh_cam0_ang40.png};
		
		\addplot [line width=1.5, color=udsgreen, semithick] table[col sep=comma, x=im0_p1, y=im0_p2] {data/ang40_img_pos.csv};
		
		\legend{,Projected ICP reconstruction };
		
		\nextgroupplot[
		title={Target Image Camera 1},
		]
		\addplot graphics [xmin=-0.5, xmax=2448, ymin=2048, ymax=-0.5] {comparison_diff_render/curve_mesh_cam1_ang40.png};
		
		\addplot [line width=1.5, color=udsgreen, semithick] table[col sep=comma, x=im1_p1, y=im1_p2] {data/ang40_img_pos.csv};
		
		\nextgroupplot[
		title=Fitted Spline -- DiffRender
		]
		\addplot graphics [xmin=-0.5, xmax=2448, ymin=2048, ymax=-0.5] {comparison_diff_render/im0_ang40.png};
		
		\nextgroupplot[
		title=Fitted Spline -- DiffRender
		]
		\addplot graphics [xmin=-0.5, xmax=2448, ymin=2048, ymax=-0.5] {comparison_diff_render/im1_ang40.png};
		
	\end{groupplot}
	
\end{tikzpicture}
	\caption{Comparison of rendered image (upper row) with rendered Bézier curve (bottom row) for the angle \SI{40}{\degree}, Left: Camera 0, Right: Camera 1. We find that the reconstruction loosely resembles the rendered image, but the initial bending of the first curve segment is not captured well, resulting in errors in subsequent segments.
	The ICP reconstruction result is projected to the target images in the upper row.}
	\label{fig:comparison_images}
\end{figure}

\begin{figure}[!htbp]
	\centering
\begin{tikzpicture}[scale=0.8]
	
	\definecolor{steelblue31119180}{RGB}{31,119,180}
	
	\begin{groupplot}[group style={group size=2 by 1},
		xticklabels={},
		yticklabels={},
		xmin=-0.5, xmax=2448,
		x dir=reverse,
		ymin=-0.5, ymax=2048,]
		\nextgroupplot[
		title={Camera 0},
		]
		
		\addplot graphics [xmin=-0.5, xmax=2448, ymin=2048, ymax=-0.5] {p_parameter_study/im0_ang60.png};
		
		\addplot [line width=1.5, color=udsgreen, semithick] table[col sep=comma, x=im0_p1, y=im0_p2] {data/p_norm_parameter_study/icp_img_pos_ang60.csv};
		
		\legend{,Projected ICP reconstruction };
		
		\nextgroupplot[
		title={Camera 1},
		]
		\addplot graphics [xmin=-0.5, xmax=2448, ymin=2048, ymax=-0.5] {p_parameter_study/im1_ang60.png};
		
		\addplot [line width=1.5, color=udsgreen, semithick] table[col sep=comma, x=im1_p1, y=im1_p2] {data/p_norm_parameter_study/icp_img_pos_ang60.csv};
		
	\end{groupplot}
	
\end{tikzpicture}
	\caption{Comparison of the \os algorithm reconstruction for the angle \SI{60}{\degree}, Left: Camera 0, Right: Camera 1. We observe that the algorithm converged to a local minimum, bending into the wrong direction in the Camera 0 image, resulting in larger errors compared to other angles.}
	\label{fig:bad_convergence}
\end{figure}

\newpage
\AtEndDocument{%
	\clearpage
	\listoffigures
}

\end{document}